\documentclass[10pt,journal,compsoc]{IEEEtran}
\usepackage{color}
\usepackage{xcolor}
\usepackage{graphicx}
\usepackage{colortbl}
\usepackage{subfigure}
\usepackage{amsmath,amssymb} 
\usepackage{multirow}
\usepackage{makecell}
\usepackage{bbding}
\usepackage{mathrsfs}
\usepackage{tabularx}
\usepackage{booktabs}
\usepackage[pagebackref=true,breaklinks=true,colorlinks,bookmarks=false]{hyperref}
\usepackage{diagbox}
\usepackage{amsthm}%
\usepackage{url}
\usepackage{tikz}
\newtheorem{theorem}{Theorem}[section]
\newtheorem{lemma}[theorem]{Lemma}

\newtheorem{proposition}[theorem]{proposition}

\begin{document}
\title{\includegraphics[scale=0.13]{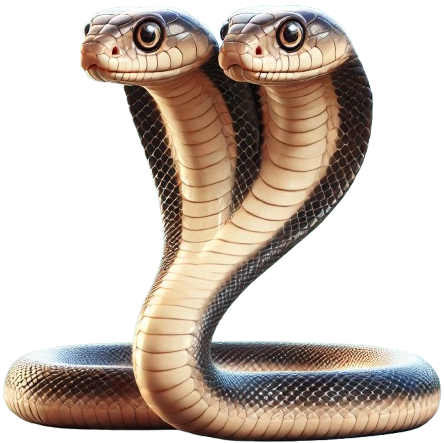}MambaX: Image Super-Resolution with State Predictive Control}

\author{Chenyu~Li,
        Danfeng~Hong,~\IEEEmembership{Senior Member,~IEEE,}
        Bing Zhang,~\IEEEmembership{Fellow,~IEEE,}
        Zhaojie~Pan,
        Naoto Yokoya,~\IEEEmembership{Member,~IEEE,}
        and Jocelyn~Chanussot,~\IEEEmembership{Fellow,~IEEE}
        
\IEEEcompsocitemizethanks{
\vspace{-5pt}
\IEEEcompsocthanksitem C. Li, D. Hong, and Z. Pan are with Southeast University, Nanjing 210096, China.
\IEEEcompsocthanksitem B. Zhang is with the Aerospace Information Research Institute, Chinese Academy of Sciences, 100094 Beijing, China, and the College of Resources and Environment, University of Chinese Academy of Sciences, Beijing 100049, China, and also with the School of Mathematics, Southeast University, 210096 Nanjing, China.
\IEEEcompsocthanksitem N. Yokoya is with the Department of Complexity Science and Engineering, Graduate School of Frontier Sciences, the University of Tokyo, Chiba 277-8561, Japan.
\IEEEcompsocthanksitem J. Chanussot is with Univ. Grenoble Alpes, Inria, CNRS, Grenoble INP, LJK, Grenoble 38000, France.
}
}


\IEEEtitleabstractindextext{%
\begin{abstract}
Image super-resolution (SR) is a critical technology for overcoming the inherent hardware limitations of sensors. However, existing approaches mainly focus on directly enhancing the final resolution, often neglecting effective control over error propagation and accumulation during intermediate stages. Recently, Mamba has emerged as a promising approach that can represent the entire reconstruction process as a state sequence with multiple nodes, allowing for intermediate intervention. Nonetheless, its fixed linear mapper is limited by a narrow receptive field and restricted flexibility, which hampers its effectiveness in fine-grained images. To address this, we created a nonlinear state predictive control model \textbf{MambaX} that maps consecutive spectral bands into a latent state space and generalizes the SR task by dynamically learning the nonlinear state parameters of control equations. Compared to existing sequence models, MambaX 1) employs dynamic state predictive control learning to approximate the nonlinear differential coefficients of state-space models; 2) introduces a novel state cross-control paradigm for multimodal SR fusion; and 3) utilizes progressive transitional learning to mitigate heterogeneity caused by domain and modality shifts. Our evaluation demonstrates the superior performance of the dynamic spectrum-state representation model in both single-image SR and multimodal fusion-based SR tasks, highlighting its substantial potential to advance spectrally generalized modeling across arbitrary dimensions and modalities. 
\end{abstract}

}

\maketitle

\IEEEdisplaynontitleabstractindextext

%
\IEEEpeerreviewmaketitle

\section{Introduction}\label{sec:introduction}
\IEEEPARstart{H}{igh-resolution} optical imaging is crucial in remote sensing \cite{LI2024CasFormer}, computer vision \cite{KXNet}, microscopy \cite{2021Multi-Modal}, and astronomy for target identification, fine-structure analysis, and scientific research \cite{ssfusion_tra_tensor}. However, optical imaging is inherently limited by diffraction, sensor sampling rate, and environmental noise, where hardware constraints impose significant costs and technical challenges on resolution enhancement. 
Super-resolution (SR) techniques overcome these limitations through computational methods, enhancing image quality without hardware modifications \cite{MHF-net,hongdanfeng2023srmapping}. 

\textbf{Task Statement.} SR can be categorized into two types according to modalities: single-image SR (SISR), which addresses the ill-posed problem in a strict sense, and multimodal image fusion-based SR (MFSR), where the integration of additional modality information serves to incorporate supplementary prior knowledge \cite{2024Spatial-Frequency}. Mathematically, SISR is an inherently challenging task to estimate the high-resolution image (HR) $\mathbf{X}\in\mathbb{R}^{H \times W \times C}$ from an observed low-resolution image (LR) $\mathbf{Y}\in\mathbb{R}^{h \times w \times C}$ that suffers severe information degradation, which can be expressed as: 
\begin{equation}
    \begin{aligned}
     \label{eq1}
        &\mathbf{Y}=\mathbf{K}\mathbf{X}+\mathcal{N},\\
    \end{aligned}
\end{equation}
where $\mathbf{K}$ represents the integrated degradation operator, which includes degradations such as blurring, downsampling, uneven intensity distribution, and missing information \cite{blindsuperreview}, $\mathcal{N}$ represents generalized noise. 

\begin{figure*}[!t]
 \centering
    \includegraphics[width=1.00\textwidth]{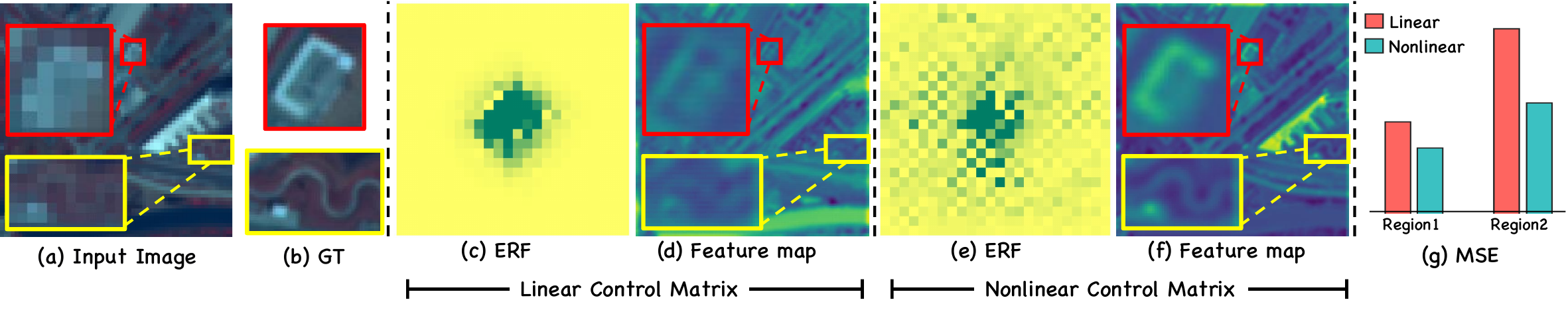}
        \caption{The Comparison of linear and nonlinear control matrix operators. (a) and (b) are the input low-resolution image and local enlargement of GT. (c) and (e) illustrate the effective receptive field at the center point of the red box, highlighting that the dynamic approach provides a broader receptive field. (d) and (f) further compare the intermediate feature maps of both methods, demonstrating that the dynamic method produces feature maps with finer details, primarily due to its larger local detail receptive field. (g) shows the mean squared error (MSE) results of two local enlargements.}\label{motivation}
\end{figure*}

In contrast to SISR, multimodal image fusion-based SR emphasizes integrating data from multiple sensors or imaging modalities as complementary references to enhance image resolution and quality. This can be expressed as:
\begin{equation}
    \begin{aligned}\label{eq2}
        &\mathbf{Y}_{\mathrm{H}} =\mathbf{X}\mathbf{B}+\mathcal{N}_{LH},\\
        &\mathbf{Y}_{\mathrm{M}} =\mathbf{C}\mathbf{X}+\mathcal{N}_{LM},\\
    \end{aligned}
\end{equation}
where $\mathbf{X} \in R^{L\times W\times H}$ is the original image that, subjected to spatial degradation $\mathbf{B}$ and affected by noise $\mathcal{N}_{LH}$, yields LR image $\mathbf{Y}_{\mathrm{LH}}$. Meanwhile, when it experiences spectral degradation $\mathbf{C}$ and is also impacted by noise $\mathcal{N}_{LM}$, it produces a low spectral resolution image $\mathbf{Y}_{\mathrm{LM}}$. 

\textbf{Technology Evolution.} Early approaches mainly leverage the intrinsic mathematical and physical characteristics of the imagery \cite{gao2022bayesian,lichengyu2024learn-prior}. Representative methodologies cover sparse \cite{2010Sparse,lichengyu2023lowrank} and dictionary learning \cite{2013Dictionaries}. The core objective of MFSR is to utilize the complementary relationship between the auxiliary image and the LR to achieve image resolution enhancement \cite{2024Multi-Modal}. Within the realm of traditional MFSR-based methods, model-based approaches consistently demonstrate the most outstanding performance \cite{ssfusion_tra_tensor}. 

Unlike conventional methods, deep learning (DL) approaches exhibit a superior ability to extract nonlinear features \cite{2023diffusion,2015CNN}. To enhance SISR performance, multiscale feature extraction, preservation of local details, global contextual modeling \cite{CNN_sup_comple}, and deep hierarchical representations \cite{2022lap} are commonly leveraged. For the MFSR task, pixel-level fusion through input stacking \cite{lag}. Transformer \cite{dosovitskiy2020vit}, with its unique attention mechanism, has emerged as a dominant paradigm in recent research. In \cite{unrolling_sup_tensor}, transformers are employed for modeling long-term temporal dependencies across spectral channels and spatial information \cite{essaformer}. Moreover, \cite{transformer_sup_token} enhances the transformer network by optimizing token selection, significantly improving the efficiency of self-attention.

\textbf{Emerging State Space Models (SSMs).} Recently, Mamba \cite{gu2024mambalineartimesequencemodeling} has demonstrated promising results in sequence modeling with SSM by linear time, which is used to describe the behavior of dynamic systems. It employs a set of first-order differential equations or difference equations to represent the evolution of internal states ${{h}}(t)'$, while another set of equations defines the relationship between the states and outputs:
\begin{equation}
    \begin{aligned}
     \label{eq3}
        &h\left ( t \right ) ^{'} = \mathbf{A}h(t)+\mathbf{B}x(t),\\
        &y(t)=\mathbf{C}h(t)+\mathbf{D}x(t),\\
    \end{aligned}
\end{equation}
where ${h}(t)$ represents the system state that is governed by the system matrix $\mathbf{A}$, which encompasses all the requisite variables to fully characterize the system condition. ${x}(t)$ represents the input state vector that captures the influence of external inputs on the system, influenced by the input control matrix $\mathbf{B}$. Similarly, in the observation equation, the control variable $\mathbf{C}$ describes the mapping of the state $h(t)$ to the output $y(t)$, and the control variable $\mathbf{D}$ represents the direct impact of the $x(t)$ on $y(t)$. To improve the operability in DL, Eq. \ref{eq3} needs to be discretized:
\begin{equation}
    \begin{aligned}
     \label{eq5}
        &h_{i}=\bar{\mathbf{A}} h_{i-1}+\bar{\mathbf{B}}x_{i},\\
    \end{aligned}
\end{equation}
\begin{equation}
    \begin{aligned}
     \label{eq6}
        &y_{i}=\mathbf{C}h_{i}+\mathbf{D}x_{i},\\
    \end{aligned}
\end{equation}
where, $h_{i-1}$ denotes prior state, and $h_{i}$ is the present state. $\bar{\mathbf{A}}$ is constructed by HiPPO \cite{gu2020hippo}, which can ensure the mathematical approximation of all previous histories. It is worth noting that to accommodate the different effects of the input characteristics on the system, Mamba modified the fixed parameters to linear transformations.

\subsection{Limitations in Linear Expression: Poor Flexibility to Fit Arbitrary Scenarios.}
Existing SSMs have demonstrated promising performance advantages, with their efficacy being largely attributed to the feature learning and expressive capacity of several key control matrices \textbf{B}, \textbf{C}, and discretization $\mathbf{\bigtriangleup}$ \cite{10812905}. While its parameters derived from a simple linear generator contribute to reduced computational time to some extent, their control capacity is evidently at odds with the inherent complexity of time-varying dynamic systems. As illustrated in Fig. \ref{motivation} (b) and (c), the control matrices obtained through linear layers exhibit a limited receptive field for feature representation, whereas dynamic non-linear computations, enhancing the adaptability, offer a significantly broader feature expressiveness.

Furthermore, the heterogeneity is not confined to the distinct modalities within the image spatial domain but also extends to the interplay between the image domain and the sequence domains. In the presence of such intricate intra-domain and inter-domain heterogeneous mappings, the approximation capacity of the linear operator is inadequate. This inherent and irreducible heterogeneity inevitably accumulates and amplifies during the training process, leading to issues such as model non-convergence or overfitting.
\subsection{Contributions of Our MambaX}
Building upon the aforementioned challenges and findings, we introduce \textbf{MambaX}, which leverages dynamic non-linear mappings to compute multistage differential coefficients of state models. Employing learnable control matrices, MambaX facilitates the linear fitting of state equations at each stage, thereby enabling the control and optimization of the SR blind process. To address the heterogeneity arising from domain discrepancies, we introduce domain-crossing transition operators that progressively mitigate error accumulation. Furthermore, we put forward the first concept, cross-control fusion for multimodal states, which cross-control matrices that carry the proprietary properties of multiple modalities to a unified state space, to realize fusion. Specifically, the contributions of this work can be summarized as follows:
\begin{itemize}
     \item We propose a nonlinear state matrix predictive control modeling framework, called \textbf{MambaX}, which is designed for dynamic, time-varying systems. MambaX enables the high-fidelity approximation of images across arbitrary resolutions and modalities. 
    
     \item In MambaX, a novel nonlinear state predictive control (nSPC) is introduced for dynamically approximating multistage differential coefficients of state-space models in a nonlinear fashion. nSPC can implicitly encode spatial and spectral attributes by leveraging a custom-designed neural network, which facilitates the dynamic learning of optimal matrix parameters for intermediate states. This enables effective control and optimization of the SR-blind process.
     
    \item A cross-domain transition learning strategy is proposed to progressively mitigate the heterogeneity arising from domain discrepancies and complex degradation processes, enabling adaptive transfer from spectral space to state space.
    
    \item A pioneering concept, \textbf{Cross-control Fusion (Cc)} is introduced, which operates on the control-state matrices rather than directly fusing state features, where matrix parameters govern the intersection and transfer of multimodal state attributes within a unified state space, enabling seamless multimodal SR fusion.  
\end{itemize}

\section{MambaX: Dynamic Nonlinear State Predictive Control}
Real-world images are inherently complex, exhibiting particularly strong nonlinear characteristics. The existing Mamba models heavily rely on linear layers to predict dynamic parameters; however, such linear expressions inevitably introduce significant discrepancies from real-world conditions. According to \cite{mathematicalssm}, nonlinear control introduces terminal constraints in the objective function and dynamic optimization, allowing it to approximate the global optimum more effectively than linear approaches \cite{Grancharova2012}. Motivated by this principle, we propose a dynamic nonlinear state predictive control (nSPC) in the designed MambaX. Besides, MambaX incorporates a progressive cross-domain mapping strategy for the intricate degradation process, enabling a more accurate feature alignment and lossless mapping between image space to state space.

\subsection{Nonlinear State Predictive Control (nSPC)}
Initially, owing to the image that may originate from disparate modality and heterogeneous spaces, the feature alignment and state transitional mapping become the critical steps for its incorporation into subsequent computations, designed to assimilate and reconcile these heterogeneous spaces, which can be expressed:
\begin{equation}
    \begin{aligned}
    \label{eq001}
        [u_1,\cdots,u_i] = \mathcal{E}(x_1,\cdots,x_i),
    \end{aligned}
\end{equation}
where $u_i$ represents the input space state variable, $x_i$ represents the input image, such as the upsampled low-resolution image and the panchromatic image. $\mathcal{E}$ achieves the assimilation transition between heterogeneous spaces of different images (this will be specifically introduced in Section \ref{Implementing}).

For the dynamic system between $u$ and $h$, we choose to use the SSM from Kalman:
$$h'(t) = Ah(t) + Bu(t), \quad y(t) = Ch(t) + Du(t),$$
Discretization is performed using zero-order hold (ZOH) \cite{gu2024mambalineartimesequencemodeling}. The discretized matrices are given by
$$\bar{A} = e^{\Delta A}, \quad \bar{B} = (\Delta A)^{-1} (e^{\Delta A} - I) \Delta B.$$
where the parameter $\Delta$ serves as the temporal scaling factor. Then, we have
$$h_{i}=\bar{A}h_{i-1}+\bar{B}u_{i}, \quad y_{i}=Ch_{i}+Du_{i},$$
where
\begin{equation}
    \begin{aligned}
    \label{eq002}
        h_i &=\bar{A}_ih_{i-1}+\bar{B}_iu_i,\\
        &=\bar{A}_{i}(\bar{A}_{i-1}h_{i-2}+\bar{B}_{i-1}u_{i-1})+\bar{B}_iu_i, \\
        &=\bar{A}_{i}(\bar{A}_{i-1}(\bar{A}_{i-2}h_{i-3}+\bar{B}_{i-2}u_{i-2})+\bar{B}_{i-1}u_{i-1})+\bar{B}_iu_i, \\
        &=\bar{A}_{i}\cdots\bar{A}_{1}h_{0}+\bar{A}_{i}\cdots\bar{A}_{2}\bar{B}_{1}u_{1}+\cdots+\bar{A}_{i}\bar{A}_{i-1}\bar{B}_{i-2}u_{i-2}  \\
        &\quad \ +\bar{A}_{i}\bar{B}_{i-1}u_{i-1}+\bar{B}_{i}u_{i}, \\
    \end{aligned}
\end{equation}
and
\begin{equation}
    \begin{aligned}
    \label{eq003}
        y_i &= C_ih_i+D_iu_i, \\
        &= C_i(\bar{A}_{i}\cdots\bar{A}_{1}h_{0}+\bar{A}_{i}\cdots\bar{A}_{2}\bar{B}_{1}u_{1}+\cdots+ \\
        &\quad \ \bar{A}_{i}\bar{A}_{i-1}\bar{B}_{i-2}u_{i-2}+\bar{A}_{i}\bar{B}_{i-1}u_{i-1}+\bar{B}_{i}u_{i}\big)+D_iu_i, \\
    \end{aligned}
\end{equation}
where $x_i$ represents the input space state variable obtained from Eq. \ref{eq001}, $h_i$ is the state modeling variable, $h_0$ is the initial state. We can see that the state variable $h_i$ is no longer only influenced by its previous state $h_{i-1}$, but instead aggregates all previously modeled states. 

Next, we introduce new variables, such as $\mathcal{A}$, to combine some terms in Eq.\ref{eq002} and Eq.\ref{eq003} as follows:
\begin{equation}
    \begin{aligned}
        h_i =\bar{\mathcal{A}}_{1:i}&h_0+\Sigma_{j=1}^{i}\bar{\mathcal{B}}_ju_{j},\quad y_i =\mathcal{C}_ih_0+\Sigma_{j=1}^{i}\mathcal{D}_ju_j,
    \end{aligned}
\end{equation}
where 
\begin{equation}\nonumber
    \begin{aligned}
     \bar{\mathcal{A}}_{j:i}:=&\begin{cases}
            \bar{A}_{i}\bar{A}_{i-1}\cdots\bar{A}_{j},\ j\leq i,\\
            1,\ j>i, 
        \end{cases}
        \bar{\mathcal{B}}_j:=\bar{\mathcal{A}}_{j+1:i}\bar{B}_{j}, \\
        \mathcal{C}_i:=&C_i\bar{\mathcal{A}}_{1:i},
        \quad \mathcal{D}_j:=\begin{cases}
            C_i\bar{\mathcal{A}}_{j+1:i}\bar{B}_{j},\ j< i,\\
            C_i\bar{B}_{i}+D_i,\ j=i.\\
            \end{cases}
    \end{aligned}
\end{equation}

We can see that the state variable $h_i$ is actually controlled by all past inputs $u_i$ starting from the initial state $h_0$, and this can also be easily observed from the structure of the matrix $\bar{\mathcal{A}}$, $\bar{\mathcal{B}}$ and $\mathcal{C}$. Therefore, the matrix $\mathcal{C}$, which controls the transition from the state to the output, plays a crucial role, and the discretization operator $\Delta$ is also important. Both have a large impact on the representational ability of the state model.

\begin{figure*}[!t]
      \centering
	   \includegraphics[width=1.0\textwidth]{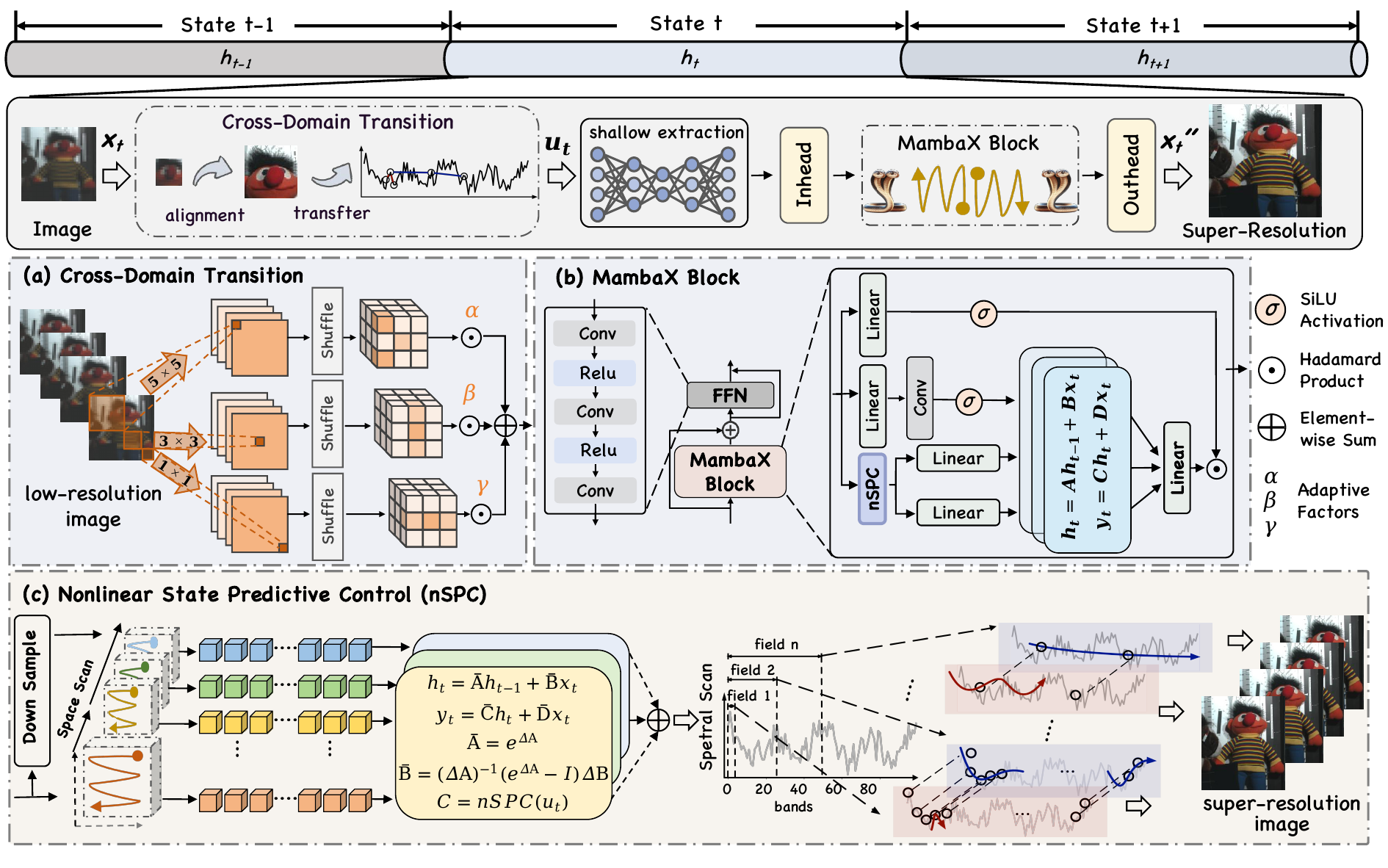}
      \caption{An illustrative workflow of the proposed MambaX, whose each state $t$ is unfolded into cross-domain transition and nonlinear state predictive control (nSPC). (a) the cross-domain transition process, highlighting degradation alignment and domain transfer. (b) provides a detailed view of the block structure in MambaX. (c) formulates the nSPC. }
\label{fig:workflow}
\end{figure*} 

Specifically, we first improve the learning of the discretization operator $\Delta$:
\begin{equation}
    \begin{aligned}
        \Delta_j^{D}:=&\Delta(x_j), \\
        =&\mathrm{softplus}\left\{\sigma[K_2(K_1x_j+b_1)+b_2]\right\}, \\
        &j\in\{1,\cdots,T\},
    \end{aligned}
\end{equation}
where $K$ and $b$ are the weights and biases learned by CNNs with appropriate dimensions. Specifically, the first convolutional layer ($K_1,b_1$) operates only in the spatial dimension, while the second layer ($K_2,b_2$) operates in the channel dimension. This approach enables fine-grained feature extraction and fusion.

For state transition mapping $\mathcal{C}$, we similarly use non-linear learning methods, yielding the following result:
\begin{equation}
    \begin{aligned}
        C_j^{D}&:=C(x_j), \\
        &=\sigma\left[K_2(K_1x_j+b_1)+b_2\right], j\in\{1,\cdots,T\}.
    \end{aligned}
\end{equation}
Up to this point, the entire process of nSPC can still be described by three equations:
\begin{equation}
    \begin{aligned}
        [u_1,\cdots,u_i] &= \mathcal{E}(x_1,\cdots,x_i), \\
        h_{i}&=\bar{A}_i^{D}h_{i-1}+\bar{B}_i^{D}u_{i}, \\ 
        y_{i}&=C_i^{D}h_{i}+Du_{i}, \\
    \end{aligned}
\end{equation}
where $\bar{A}_i^{D} = e^{\Delta_i^{D} A_i}$, $\bar{B}_i^{D} = (\Delta_i^{D} A_i)^{-1}$ $(e^{\Delta_i^{D} A_i} - I) \Delta_i^{D} B_i$. Above, we have reconstructed the learning of the state transition matrix and discretization operator in our nSPC model, where the state variable $h_i$ acquires richer context, and the output $y_i$ extracts more complex features, thereby improving adaptability and providing a more comprehensive understanding of the input. Some specific details will be elaborated in the following sections. 

\subsubsection{Convergence Analysis}
The following part will analyze the convergence of nSPC, which gives a mathematically rigorous guarantee of its feasibility. According to \cite{ssmlayerwisenonlinearity}, we have
\begin{lemma}
For any given continuous function $f$ over a compact set $K$. Then there exists a sequence of state-space models that can approximate the sequence relationship $f:(u_1,\cdots,u_T)\to(f(u_1),\cdots,f(u_T))$.
$$\sup_t\sup_x|f(u)-\hat{y}|\leq\epsilon,$$
\end{lemma}
\noindent and the model is constructed by
\begin{equation}
    \begin{aligned}
        h_{i}&=\bar{A}_i^{D}h_{i-1}+\bar{B}_i^{D}u_{i}, \\ 
        y_{i}&=C_i^{D}h_{i}+Du_{i}. \\
       \end{aligned}
\end{equation}
Next, we introduce the Weierstrass Approximation Theorem:
\begin{theorem}
    \textbf{(Weierstrass Approximation Theorem)}: Let $f$ be a continuous function on $\mathbb{R}$. Then, there exists a sequence of polynomial functions $\{P_n(x)\}$, such that:
    $$P_n(x) \xrightarrow{uniformly} f(x) \;\; \text{on} \;\; \mathbb{R}.$$
    This implies that for any $\epsilon > 0$, there exists a polynomial $P_n(x)$ such that:
    $$|f(x) - P_n(x)| < \epsilon, \quad \text{for all} x \in \mathbb{R}.$$
\end{theorem}
Combining the Lemma and the Theorem, we can easily conclude that
\begin{proposition}
For any bounded causal continuous sequence to sequence relationship $H:\{x_i\}_{i=1}^T\to\{y_i\}_{i=1}^T$ and tolerance $\epsilon>0$, there exists a hidden dimension $m$ and corresponding state-space model such that the error of approximation
$$\sup_t\sup_x|y_i-\hat{y}_i|\leq\epsilon,i\in\{1,\cdots,T\}.$$
\end{proposition}
Above, we describe a brief convergence analysis.
$\hfill \blacksquare$

\subsection{Implementing nSPC in MambaX}\label{Implementing}
 
\subsubsection{Progressive Cross-Domain Transition}
Current SR techniques predominantly employ fixed convolutional operations or pixel-shuffle mechanisms for upsampling purposes. Nevertheless, the degradation phenomenon exhibits substantial complexity, rendering conventional upsampling strategies insufficient for capturing full characteristics and struggling to handle the significant heterogeneity between cross-domains. To connect the original high-dimensional observations to the latent state space, we introduce a transfer operator with an adaptive approach to simulate the degradation blur kernel. This kernel implements the gradual mapping from the original space to state space within non-linear degeneration.

Specifically, we first obtain different feeling fields $X_{i,b_{i}}$ using various mapping operators:
\begin{equation}
    \begin{aligned}
     \label{eq26}
        X_{i,b_{i}}=\mathcal{F}_\mathrm{PS} (\mathrm{ConV}(X_{i-1,c})), 
    \end{aligned}
\end{equation}
where $\mathcal{F}_\mathrm{PS}(\cdot)$ is the pixelshuffle function, and $X_{i-1,c}$ is the output of the $i-1$-th layer. 

Subsequently, to facilitate a more flexible transformation from image space into state space domain, we introduce learning weighting factors that adaptively fuse the three bases. This  mechanism can be formally expressed as:
\begin{equation}
    \begin{aligned}
     \label{eq29}
        X_{i,c}=\sum_{i=1}^{3} (\mathrm{softmax} ([\alpha, \beta, \gamma])_{i} \odot   X_{i,b_{i}}),
    \end{aligned}
\end{equation}
where $\alpha$, $\beta$ and $\gamma$ are adaptive weights. Moreover, the adaptive weights for each layer are optimized independently during training, greatly enhancing the model’s ability to capture complex degradation patterns and blur kernels.

\subsubsection{Dynamic Discretization in Nonlinear System}
MambaX introduces two cascaded fully-connected networks to dynamically distill pixel-wise channel features while strategically integrating dynamic structure-channel-aware modules between adjacent fully-connected layers for adaptive $\Delta$ generation. Assuming the input features to the nSPC are denoted as $X_{i}\in R^{H\times W\times c_m }$, where $H,W,C$ represent the height, width, and number of bands of the input image, respectively. This architecture can be formally expressed through the following mathematical formulation:
\begin{equation}
    \begin{aligned}
        \mathrm{DSC} _{\Delta}(X_{i})=W^{\Delta}_{c_r\times c_d} (W^{\Delta}_{c_d\times c_r} (C_{l}(S_{l}(W_{c_m\times c_d}X_{i})))) 
    \end{aligned}
\end{equation}
\begin{equation} 
     \mathrm{s.t.} \\
    \left\{\begin{aligned} 
        &S_{l}(\cdot )=\mathrm{BN} (\mathrm{Con} _{7\times 7,d}(\mathrm{Con} _{5\times 5,d}(\cdot)))\\
        &C_{l}(\cdot) = \mathcal{F} _{c,3}\Bigl(\mathcal{F}_{c,2}\bigl(\mathcal{F} _{c,1}(X_{i,S}, S_{i,l}\bigr), C^{i}_{spa}\Bigr), X_{i,S}\Bigr),\\
    \end{aligned}\right.   
\end{equation} 
where $W_{c_m\times c_d}X_{i}$ with channel $c_d$ is used to generate the control variable. In addition, $W^{\Delta}_{c_d\times c_r}$ and $W^{\Delta}_{c_r\times c_d}$ are sequentially applied to compress and expand the feature dimensions, thereby preserving critical information, where $c_d \gg  c_r$.

\subsubsection{Generalized Learning of Control Matrix}
Mamba generates $C$ using only a single linear layer, limiting its ability to model complex scenarios. Hence, we integrate a dynamic non-linear operator into $C$'s generation:
\begin{equation}
    \begin{aligned}
     \label{eq31}
        \mathrm{DSC} _{C} (S_{i})=W^{C}_{c_d\times c_s} (C_{l}(S_{l}(W_{c_m\times c_d}X_{i}))),
    \end{aligned}
\end{equation}
where the linear layer $W^{C}_{c_d\times c_s}$ projects $C$ to ensure its dimensionality matches that of the state-space representation.

As evidenced by the preceding analysis, distinct generation mechanisms are employed for different state control matrices to enhance the dynamic properties of the state-space equations. Given that matrix $A$ and $B$ govern state transition dynamics and input coupling, respectively, we implement indirect modulation through the dynamic generation of $\Delta$ to preserve the stability of the state-space model. In contrast, matrix $C$, which regulates state output projection, undergoes direct dynamic generation. This architectural strategy equips MambaX with an enhanced capability to adapt to intricate structural variations.

\subsection{Cross-Control Fusion for Multimodal States}
To enable the auxiliary image to effectively guide the reconstruction of spatial information in the LR image, we generate the key control variables $B$ using the auxiliary image. The matrix $A$ continues to use its original initialization method to ensure the stability of the state process. Additionally, considering the different roles of $B$, $C$, and $\bigtriangleup$ in the state and output equations, we employ distinct learning methods to form these control variables. This process can be depicted as
\begin{equation}
    \begin{aligned}
     \label{eq16}
        \bar{B}^{i} =\mathrm{D} _{B^{i}}(A^{i},\mathrm{Linear}_B(S_{i}),\bigtriangleup),
    \end{aligned}
\end{equation}
where $C^{i} =\mathrm{DSC} _{C^{i}} (X_{f,s}^{i})$ and $\triangle^{i} =\mathrm{DSC} _{\triangle^{i} } (X_{f,s}^{i})$ represent the dynamic learnable operators of $C$ and $\bigtriangleup$, respectively. The simple generation method for $B$ stabilizes $\bar{B}$, thereby deliberately constraining the fused state's modeling space and ensuring overall stability. 

Before formally using the state space model to fuse $X_{f,s}^{i}$ and $S_{i}$, it is necessary to Conert the control variables and $X_{f,s}^{i}$ into sequence form $\bar{B}^{i}=\left [ \bar{B}^{i}_{1},\bar{B}^{i}_{2},\cdots \bar{B}^{i}_{k},\cdots \right ]$, $ C^{i}=\left [ C^{i}_{1},C^{i}_{2},\cdots C^{i}_{k},\cdots \right ]$, $\bar{A}^{i}=\left [ \bar{A}^{i}_{1},\bar{A}^{i}_{2},\cdots \bar{A}^{i}_{k},\cdots \right ]$, and $X_{f,s}^{i}=\left [ x^{i}_{1},x^{i}_{2},\cdots x^{i}_{k},\cdots \right ]$.

Subsequently, the fusion process at each position can be represented as:
\begin{equation}
    \begin{aligned}
     \label{eq22}
        \begin{bmatrix}
h_{k}^{s} \\
y^{i}_{k}
\end{bmatrix} &=
\begin{bmatrix}
\bar{A}^{i}_{k} & \bar{B}^{i}_{k} \\
\bar{C}^{i}_{k}\bar{A}^{i}_{k} & \bar{C}^{i}_{k}\bar{B}^{i}_{k} + \bar{D}^{i}_{k}
\end{bmatrix}
\begin{bmatrix}
h_{k-1}^{s} \\
x^{i}_{k}
\end{bmatrix},\\
& =\begin{bmatrix}
\bar{A}^{i}_{k} & \mathcal{F}_{\bar{B}^{i}_{k}}(S_{i},X_{f,s}^{i})  \\
\mathcal{F}_{\bar{C}^{i}_{k}}(X_{f,s}^{i})\bar{A}^{i}_{k} & \mathcal{F}_{\bar{C}^{i}_{k}\bar{B}^{i}_{k}}(S_{i},X_{f,s}^{i}) + \bar{D}^{i}_{k}
\end{bmatrix}
\begin{bmatrix}
h_{k-1}^{s} \\
x^{i}_{k}
\end{bmatrix},\\
    \end{aligned}
\end{equation}
where ${h} _{k}^{s}$ represents the intermediate state, which also indicates the initial fusion result of the supplementary information $S_{i}$ and $X_{f,s}^{i}$. The $y^{i}_{k}$ denotes the refined fusion result, whereas in the original image $X_{f,s}^{i}$ is further integrated with the initial fusion outcome to calibrate the final fusion output. Subsequently, the outputs of state space equations are merged and rearranged to obtain $Y_{i}$.

\begin{figure*}[!t]
 	  \centering
 			\includegraphics[width=1.0\textwidth]{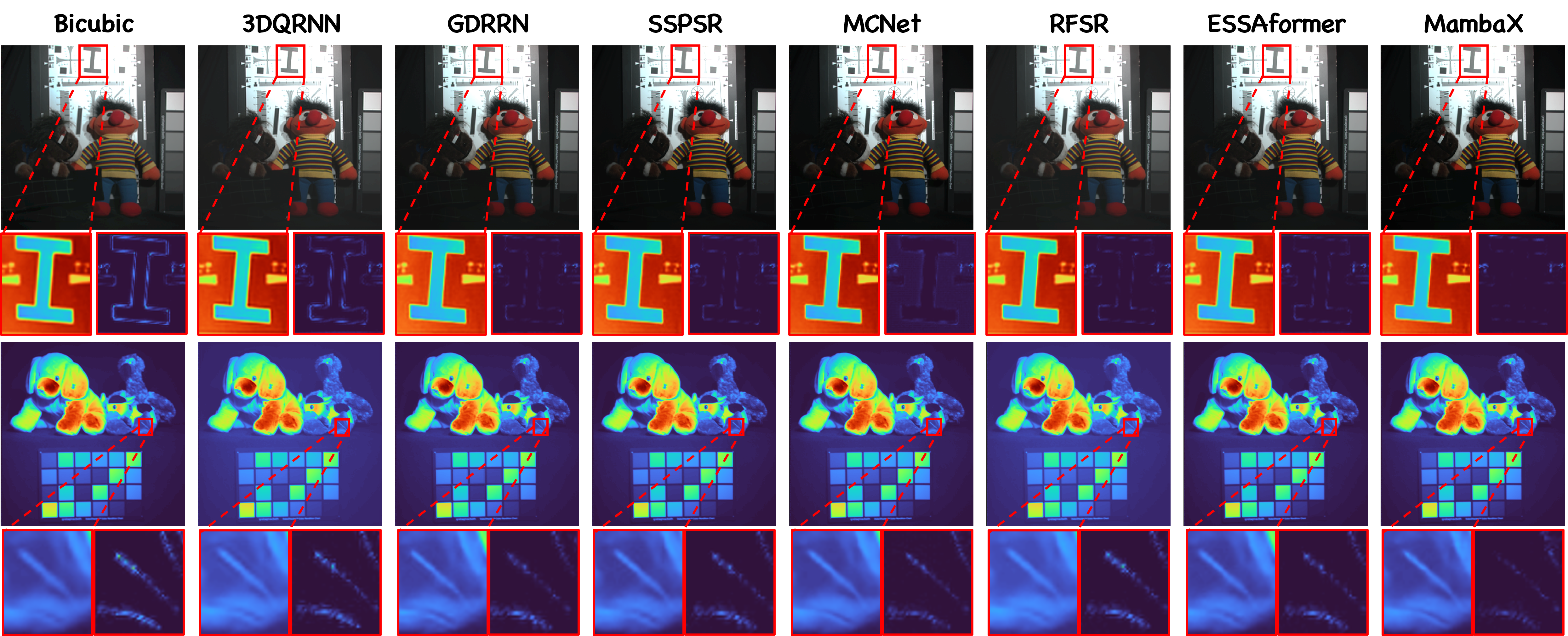}
         \caption{{Visual quality comparison on CAVE dataset. The first row shows the pseudo-color image (50,40,30) of the 8x SR result. The second row shows the MSE between the 8x SR result and the ground truth. The last row shows the local enlargement of the MSE. The bar chart on the right displays the RMSE values for each algorithm.}}
 \label{fig:cave}
 \end{figure*}
\begin{table}[!ht]
\begin{center}
\caption{Quantitative performance on testing samples of the CAVE dataset at different scale factors. \colorbox {gray!20}{\textbf{Bold}} represents the best result and {\underline{underline}} represents the second best.}
\scalebox{0.9}{
\begin{tabular}{cccccccc}
        \toprule
        {Comparison} &\multirow{3}{*}{Scale}&\multicolumn{5}{c}{CAVE}\\
        \cmidrule(lr){3-7}
        {Method} & & PSNR↑ & SSIM↑ & SAM↓ & ERGAS↓ & CC↑\\ 
        \midrule
        \midrule
        Bicubic  & \multirow{8}{*}{x2} & 39.81& 0.9658 & - & 2.59 &0.9926 \\
        3DQRNN  &  & 40.56 & 0.9694 & 1.37 &2.36 &0.9940\\
        GDRRN &  & 41.84 & 0.9726 & 1.49 & 2.12 &0.9950\\
        SSPSR &  & \underline{42.57} &\underline{0.9742} &1.25 & \underline{1.98} &0.9953\\
        MCNet &  & 42.55 & 0.9726 & 4.42 & 2.46 &\cellcolor{gray!30}\textbf{0.9955}\\
        RFSR &  & 42.24 & 0.9730 &\underline{1.18} &{2.06} &0.9948\\
        ESSAformer  &  & {42.40} & \underline{0.9742} & 1.31 & {2.00} & 0.9953 \\
        \includegraphics[scale=0.043]{fig/logo.png} MambaX &  & \cellcolor{gray!30}\textbf{43.35} &\cellcolor{gray!30}\textbf{0.9755} & \cellcolor{gray!30}\textbf{1.10} & \cellcolor{gray!30}\textbf{1.87} & \cellcolor{gray!30}\textbf{0.9955}\\
        \midrule
        Bicubic  & \multirow{8}{*}{x4} & 34.55 & 0.9151 & - & 4.50 &0.9784\\
        3DQRNN  &  & 35.72 & 0.9246 & 2.39 & 3.97 & 0.9822\\
        GDRRN  &  & 36.18 & 0.9295 & 2.16 &3.93  &0.9826\\
        SSPSR  &  & {37.25} & \underline{0.9378} & {2.01} & \underline{3.45} &\cellcolor{gray!30}\textbf{0.9848} \\
        MCNet &  & \underline{37.27} & 0.9373& \underline{1.81} & 3.47 & 0.9845\\
        RFSR  &  & 36.93 & 0.9343 & 1.62 & 3.59 &0.9837\\
        ESSAformer &  & {37.21} & {0.9373} & 1.81 & 3.47 &\cellcolor{gray!30}\textbf{0.9848}\\
        \includegraphics[scale=0.043]{fig/logo.png} MambaX &  & \cellcolor{gray!30}\textbf{37.39}  & \cellcolor{gray!30}\textbf{0.9391} & \cellcolor{gray!30}\textbf{1.56} & \cellcolor{gray!30}\textbf{3.46} &0.9845 \\
        \midrule
        Bicubic  & \multirow{8}{*}{x8} & 30.48 & 0.8469 & 2.44 &6.92 &0.9555\\
        3DQRNN  &  & 31.61 & 0.8600 & 3.07 &6.13 &0.9626\\
        GDRRN  &  & 31.69 & 0.8623 & 2.99 & 6.17 &0.9622\\
        SSPSR  &  & \underline{32.58} & \underline{0.8775}& {2.68} & {5.58} &\cellcolor{gray!30}\textbf{0.9667} \\ 
        MCNet &  & 32.64 & 0.8768 & 2.47 & \cellcolor{gray!30}\textbf{5.54} & 0.9664\\
        RFSR  &  & {32.63} & {0.8768} & \cellcolor{gray!30}\textbf{2.20} & \underline{5.55} &0.9666\\
        ESSAformer &  & {32.48} & {0.8745} & 3.80 & 5.57 &\cellcolor{gray!30}\textbf{0.9667}\\
       \includegraphics[scale=0.043]{fig/logo.png}  MambaX  &  & \cellcolor{gray!30}\textbf{32.73} & \cellcolor{gray!30}\textbf{0.8783} & \underline{2.41} & {5.57} & {0.9646}\\
        \bottomrule
    \end{tabular}}\label{tab:cave}
\end{center}
\end{table}

Finally, to prevent the model from forgetting spectral information, the fused result is further guided by $X_{f,s}^{i}$. It can be represented as
\begin{equation}
    \begin{aligned}
     \label{eq24}
        X_{i,S}=\mathrm{Linear} (\mathrm{LN} (Y_{i})\odot \mathrm{SiLU} (\mathrm{Linear} (X_{f}^{i})) ), 
    \end{aligned}
\end{equation}
where $X_{i,S}$ represents the output of the dynamic state control state space model, and $\mathrm{SiLU} (\cdot)$ represents the activation function operator (sigmoid linear unit). $\mathrm{LN} (\cdot)$ is the layernorm operator.

\section{Experiments}
In this section, we rigorously assess the performance of MambaX in three tasks: single-image computer vision image SR, single remote sensing image SR, and multimodal image fusion-based SR. 
\subsection{Experimental Setting}
\textbf{Implementation.} For the two SISR tasks, both qualitative and quantitative evaluations are conducted at degradation scales of 2, 4, and 8. In the IFSR task, we systematically assess the model’s performance on images encompassing a diverse array of spectral bands and scene types, all subjected to a 4-fold degradation factor. Furthermore, we perform an exhaustive analysis of the novel architectural components integrated into MambaX, thereby delineating the unique contributions of each module across the evaluated tasks. For all tasks, MambaX is trained using the Adam optimizer \cite{adam} with $\beta_1 = 0.9$ and $\beta_2 = 0.999$, and all experiments are implemented using the PyTorch framework on eight NVIDIA GeForce GTX L40 GPUs (48 GB each).

\noindent\textbf{Metrics.} To evaluate the comprehensive maintenance ability of comparison methods in terms of spatial consistency and spectral fidelity, four metrics were used in experiments: Peak Signal-to-Noise Ratio (PSNR), Structural Similarity Index (SSIM), Spectral Angle Mapper (SAM), and the Relative Global Dimensional Synthesis Error (ERGAS). In addition, we also applied the correlation coefficient (CC) as an evaluation metric in the SISR experiments.

\begin{table*}[!t]
\begin{center}
\caption{Quantitative performance on testing samples of the Chikusei and Pavia datasets at different scale factors. \colorbox {gray!20}{\textbf{Bold}} represents the best result and \underline{underline} represents the second best.}
\scalebox{1.0}{
\begin{tabular}{cccccccccccc}
        \toprule[1.0pt]
        \multirow{3}{*}{Comparison Methods} &\multirow{3}{*}{Scale}&\multicolumn{5}{c}{Chikusei} &\multicolumn{5}{c}{Pavia} \\
        \cmidrule(lr){3-7}\cmidrule(lr){8-12}
         & & PSNR↑ & SSIM↑ & SAM↓ & ERGAS↓ & CC↑ & PSNR↑ & SSIM↑ & SAM↓ & ERGAS↓  & CC↑\\
        \midrule\midrule
        Bicubic  & \multirow{8}{*}{x2} & 43.00& 0.9691 & 1.94 & 3.66 &0.9724 & 32.57 & 0.9116& 4.10 & 4.23 &0.9568 \\
        3DQRNN  &  & 44.74 & 0.9798 & 1.81 &3.07 &0.9803 & 33.00 &0.9227 &5.04 &4.06 &0.9596\\
        GDRRN &  & 45.14 & 0.9824 & 1.81 & 3.04 &0.9816 & 33.09 & 0.9228 &4.93 & 4.03 &0.9602\\
        SSPSR &  & 46.15 &0.9850 &1.56 & 2.67 &0.9849 &\underline{34.82} & \underline{0.9452} & 3.94 &\underline{3.33} &\underline{0.9721}\\
        MCNet &  & 46.40 & 0.9853 & 1.49 & 2.64 &\underline{0.9859}& 34.56 & 0.9428 & 3.97 & 3.43 &0.9711\\
        RFSR &  & 46.03 & 0.9847 &\underline{1.48} &\underline{{2.69}} &0.9846 &34.50 &0.9434 &\underline{3.74} &3.43 &0.9707\\
        ESSAformer  &  & \underline{46.24} & \underline{0.9854} & 1.59 & 2.70 & 0.9848 & 34.17 & 0.9369 & 4.23 &3.57 &0.9681\\
        \includegraphics[scale=0.043]{fig/logo.png} MambaX &  & \cellcolor{gray!30}\textbf{46.70} &\cellcolor{gray!30}\textbf{0.9867} & \cellcolor{gray!30}\textbf{1.43} & \cellcolor{gray!30}\textbf{2.57} & \cellcolor{gray!30}\textbf{0.9861} & \cellcolor{gray!30}\textbf{35.43} & \cellcolor{gray!30}\textbf{0.9516} & \cellcolor{gray!30}\textbf{3.64} &\cellcolor{gray!30}\textbf{3.11} &\cellcolor{gray!30}\textbf{0.9756} \\
        \midrule
        Bicubic  & \multirow{8}{*}{x4} & 37.51 & 0.8857 & 3.71 & 6.82 &0.9011 & 28.06 & 0.7260 &5.82 &7.02 &0.8776\\
        3DQRNN  &  & 38.28 & 0.9058 & 3.63 & 6.29 &0.9156& 28.43 &0.7593 &6.46 &6.74 &0.8867 \\
        GDRRN  &  & 38.91 & 0.9183 & 3.32 &5.88  &0.9259& 28.47 & 0.7607 & 6.62 &6.71 &0.8872\\
        SSPSR  &  & \underline{39.61} & \underline{0.9287} & \underline{2.81} & \underline{5.39} &\underline{0.9358} &29.11 &0.7901  &5.61 &6.22 &0.9031\\
        MCNet &  & 39.25 &0.9214& 3.10 & 5.60 &0.9318&\underline{29.12} & \underline{0.7912} & 5.84 &\underline{6.21} &\underline{0.9033}\\
        RFSR  &  & 39.13 & 0.9210 & 3.00 & 5.68 &0.9296& 29.06 & 0.7880 &\underline{5.50} &6.26 &0.9018\\
        ESSAformer &  & {39.32} & {0.9252} & 3.06 & 5.65 &0.9308& 28.82 & 0.7779 & 6.00 & 6.45 &0.8962\\
        \includegraphics[scale=0.043]{fig/logo.png} MambaX &  & \cellcolor{gray!30}\textbf{39.76}  & \cellcolor{gray!30}\textbf{0.9297} & \cellcolor{gray!30}\textbf{2.79} & \cellcolor{gray!30}\textbf{5.38} &\cellcolor{gray!30}\textbf{0.9368} & \cellcolor{gray!30}\textbf{29.28} &\cellcolor{gray!30}\textbf{0.7994} & \cellcolor{gray!30}\textbf{5.46} &\cellcolor{gray!30}\textbf{6.14} &\cellcolor{gray!30}\textbf{0.9058}\\
        \midrule 
        Bicubic  & \multirow{8}{*}{x8} & 34.46 & 0.7896 & 5.46 &9.62 &0.7899 & 25.36 & 0.5161 & \underline{7.53} &9.56 &0.7582\\
        3DQRNN  &  & 34.74 & 0.7979 & 5.48 & 9.33&0.8027 & 25.56 &0.5417 & 7.95 &9.34 &0.7685\\
        GDRRN  &  & 35.13 & 0.8165 & 5.02 & 8.95 &0.8192 & 25.47 &0.5346 & 8.04 & 9.44 &0.7625\\
        SSPSR  &  & \underline{35.43} & \underline{0.8255}& {4.80} & \underline{8.59} &\underline{0.8332} &\underline{25.60} & \underline{0.5464} & 7.92 & \underline{9.30} &0.7700\\ 
        MCNet &  & 35.20 & 0.8153 & 5.04 & 8.80 &0.8235& \underline{25.60} &0.5444 & 7.95 &\underline{9.30} &\underline{0.7723}\\
        RFSR  &  & 35.28 & 0.8189 & \underline{4.76} & 8.73 &0.8256& 25.58 & 0.5371 &7.57 &9.32 &0.7685\\
        ESSAformer &  & {35.13} & {0.8188} & 4.99 & 8.97 &0.8194 &25.54 & 0.5432 & 8.19 &9.36 &0.7664\\
        \includegraphics[scale=0.043]{fig/logo.png} MambaX &  & \cellcolor{gray!30}\textbf{35.69} & \cellcolor{gray!30}\textbf{0.8300} & \cellcolor{gray!30}\textbf{4.59} & \cellcolor{gray!30}\textbf{8.54} & \cellcolor{gray!30}\textbf{0.8335} &\cellcolor{gray!30}\textbf{25.87} &\cellcolor{gray!30}\textbf{0.5763} &\cellcolor{gray!30}\textbf{7.12} & \cellcolor{gray!30}\textbf{9.02} &\cellcolor{gray!30}\textbf{0.7863}\\
        \bottomrule[1.0pt]
    \end{tabular}}\label{Chikusei_Pavia}
\end{center}
\end{table*}
\begin{figure*}[!t]
 	  \centering
 			\includegraphics[width=1.0\textwidth]{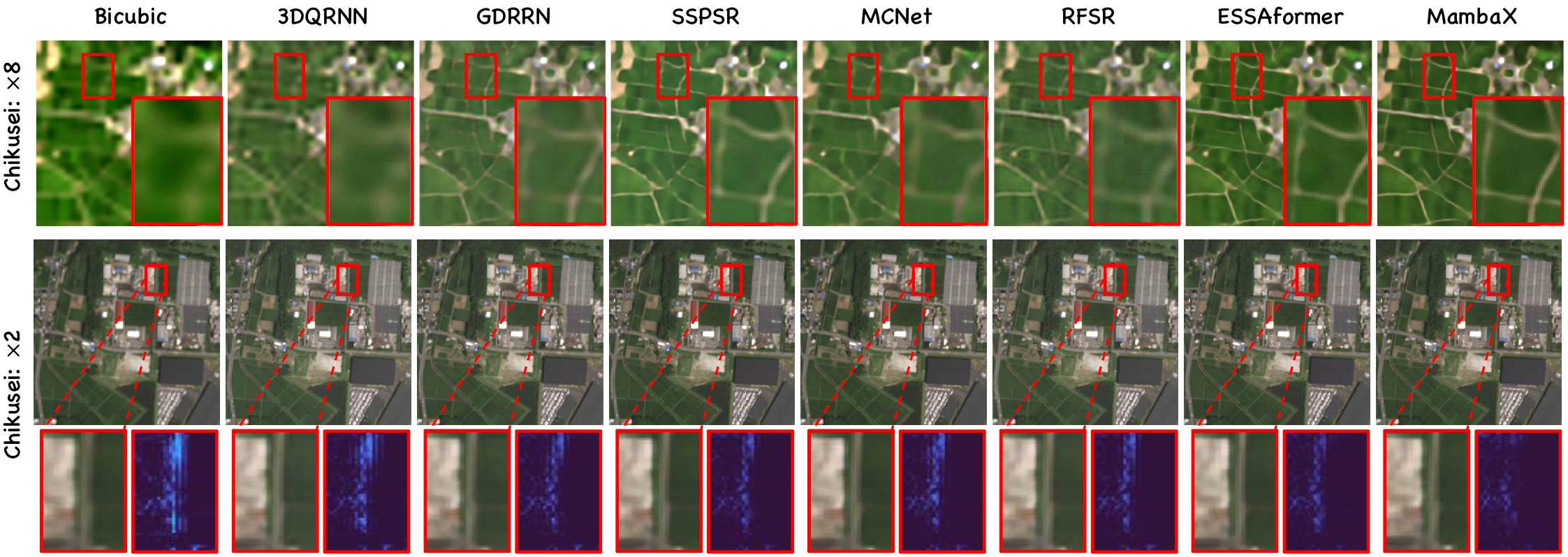}
         \caption{{Visual quality comparison on Chikusei dataset. The first and second rows show the pseudo-color image (50,40,30) of the x8 and x2 SR results, respectively. The third row shows the local enlargement of mean squared error (MSE) results between the x2 SR results and ground truth.}}
 \label{chikusei}
 \end{figure*}

\subsection{Computer Vision Image with Low-dimensional Spectral Bands} 
\subsubsection{Experimental Datasets}
\textbf{CAVE}\footnote{http://www.cs.columbia.edu/CAVE/databases/} collected by the Computer Vision Laboratory at Columbia University using an adjustable filter camera system, comprises hyperspectral images with a spatial resolution of 512×512 pixels. The dataset consists of 31 spectral bands, covering a wavelength range from 400nm to 700nm. To ensure the fairness of the experimental setup, 10 scenes were randomly selected for testing, while the remaining images were designated for training.

\subsubsection{Comparison with Professional Methods}
We compare the performance of our proposed method with several state-of-the-art image SR approaches, including 3D Conolution-based bidirectional 3D quasi-recurrent neural network (3DQRNN) \cite{3dqrnn},  mixed Conolution-based MCNet \cite{MCNET}, residual networks-based grouped deep recursive residual network (GDRRN) \cite{GDRRN}, spatial-spectral prior network (SSPSR) \cite{sspsr} and recurrent feedback network (RFSR) \cite{rfsr}, which use spectral-spatial priors, and ESSAformer \cite{essaformer}, which is based on an improved transformer. Bicubic interpolation is also used as the baseline method for comparison.

  \begin{figure*}[!t]
 	  \centering
			\includegraphics[width=1.0\textwidth]{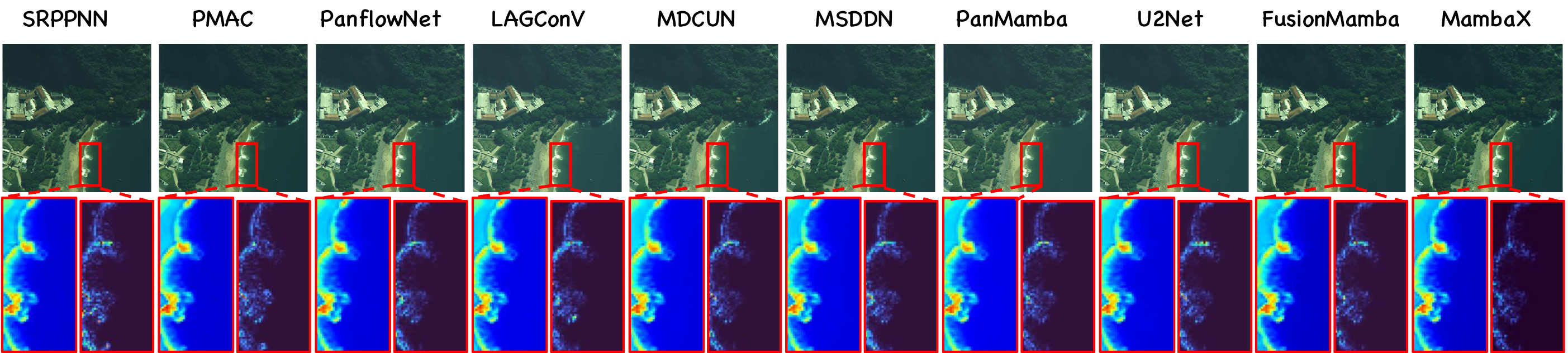}
         \caption{{Visual quality comparison on WV3 dataset. The first row shows the true color images of different results, and the second row shows the local enlargement of the mean squared error (MSE) results and the fourth channel result.}}
 \label{wv}
 \end{figure*}
 \begin{figure*}[!t]
 	  \centering
			\includegraphics[width=1.0\textwidth]{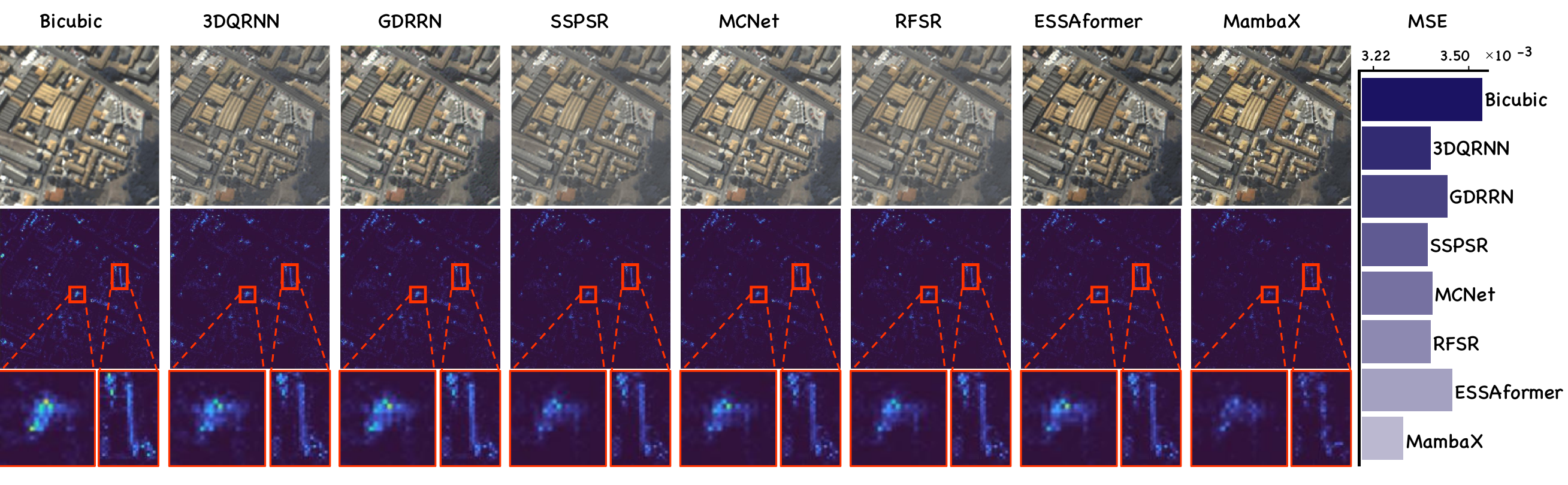}
         \caption{{Visual quality comparison on Pavia datasets. The first row shows the pseudo-color image (50,40,30) of the x2 SR results. The second row shows the local enlargement of the mean squared error (MSE) results. The bar chart on the right displays the root mean squared error (RMSE) values for each algorithm.}}
 \label{pavia}
 \end{figure*}

\noindent\textbf{Quantitative Comparison.} 
Tab. \ref{tab:cave} presents the results of various SR methods on the CAVE dataset. MambaX achieves near-optimal or even the best performance on almost all metrics under ×2 and ×4 degradation, indicating its strong capability to reconstruct spectral and spatial information at lower magnification factors. When the degradation factor increases to ×8, SSPSR and RFSR also exhibit favorable results; however, MambaX still maintains superiority in terms of PSNR and SSIM. These findings suggest that MambaX, benefiting from a global receptive field, outperforms conventional CNN-based and transformer-based models under high-degradation scenarios.

\noindent\textbf{Visual Comparison.} As shown in Fig. \ref{fig:cave}, we conduct a visual comparative analysis of various SR methods on the CAVE dataset. It can be observed that ESSAFormer and MambaX consistently maintain lower MSE values across most regions. Notably, MambaX achieves the most favorable results in preserving fine edges and texture details. Fig. \ref{figall} further presents the RMSE of MambaX across different spectral bands, demonstrating that our method achieves superior reconstruction accuracy across all bands. In summary, MambaX attains the best performance in both spatial detail reconstruction and global spectral modeling, effectively showcasing its strong potential in the field of computer vision image reconstruction.

\subsection{Single Remote Sensing Image Super-Resolution}
\subsubsection{Experimental Datasets.} 
\textbf {Chikusei}\footnote{https://naotoyokoya.com/Download.html} was captured in Chikusei, Ibaraki, using the Headwall Hyperspec-VNIR-C sensor, which has a spatial resolution of 2.5 meters with original dimensions of 2517 × 640 pixels and consists of 128 spectral bands spanning 363-1018 nm. The dataset is divided into 16 non-overlapping patches (512 × 512 × 128), with 12 used for training and 4 for testing. 

\noindent\textbf {Pavia}\footnote{https://www.ehu.eus/ccwintco/index.php/Hyperspectral\\\_Remote\_Sensing\_Scenes} is acquired by the ROSIS sensor, encompassing 115 spectral bands spanning wavelengths from 430 to 860nm. The dataset is subdivided into twelve non-overlapping patches, each of size 224 × 224 × 102. Of these, nine patches are allocated for training purposes, while the remaining three patches are designated for testing.

\subsubsection{Comparison with Professional Methods.} 
We compare the performance of our proposed method with several state-of-the-art image SR approaches, including 3D Conolution-based bidirectional 3D quasi-recurrent neural network (3DQRNN) \cite{3dqrnn}, mixed Conolution-based MCNet \cite{MCNET}, residual networks-based grouped deep recursive residual network (GDRRN) \cite{GDRRN}, spatial-spectral prior network (SSPSR) \cite{sspsr} and recurrent feedback network (RFSR) \cite{rfsr}, which use spectral-spatial priors, and ESSAformer \cite{essaformer}, which is based on an improved transformer. Bicubic interpolation is also used as the baseline method for comparison.

\begin{table*}[!t]
\begin{center}
\caption{Quantitative performance on testing samples of the Worldview III and Gaofen II datasets. \colorbox {gray!20}{\textbf{Bold}} represents the best result and \underline{underline} represents the second best.}
\scalebox{1.1}{
\begin{tabular}{ccccccccc}
        \toprule
        \multirow{3}{*}{Comparison Method} 
        & \multicolumn{4}{c}{Worldview III} &\multicolumn{4}{c}{Gaofen II} \\
        \cmidrule(lr){2-5}\cmidrule(lr){6-9}
         & PSNR↑ & SSIM↑ & SAM↓ & ERGAS↓ & PSNR↑ & SSIM↑ & SAM↓ & ERGAS↓ \\
        \midrule
        \midrule
        SRPPNN 
        & 36.53 & 0.9673 & 4.32 &2.42  & 41.81 &0.9758 &1.58 &0.7629\\
        PMAC 
        & 36.76 & 0.9686 & 4.22 & 2.36 & 39.96 & 0.9661 &1.99 & 0.9523\\
        PanflowNet 
        & 36.82 &0.9695 &4.16 & 2.34 &41.66  & 0.9759 & 1.60 &0.7667 \\
        LAGNet 
        & 36.96 & 0.9697 & 4.13 & 2.30 & 41.86 & 0.9766 & 1.58 & 0.7568 \\
        MDCUN 
        & 36.97 & 0.9707 &4.06 &2.29 &41.68 &0.9758 &1.61 &0.7742 \\
        MSDDN 
        & 37.07 & 0.9707 & 4.06 & 2.29 & \underline{42.92} & 0.9810 & \underline{1.39} &\underline{0.6673}\\
        U2Net 
        & 37.30 & 0.9726 & 3.91 & 2.20 & 42.85 & 0.9808 & 1.40 & 0.6719 \\
        PanMamba 
        & 37.28 & 0.9727 &3.93 &2.21 &42.88 &\underline{0.9813} &1.40 &0.6719 \\
        FusionMamba 
        & \underline{37.40} & \underline{0.9731} & \underline{3.87} & \underline{2.17} & 42.73 & 0.9805 & 1.42 &0.6819\\
        \midrule
        \includegraphics[scale=0.047]{fig/logo.png} \textbf{MambaX}
        & \cellcolor{gray!25}{\textbf{37.87}} & \cellcolor{gray!25}{\textbf{0.9755}} & \cellcolor{gray!25} {\textbf{3.66}} &\cellcolor{gray!25}{{\textbf{2.06}}} & \cellcolor{gray!25} {\textbf{43.65}} &\cellcolor{gray!25}{\textbf{0.9838}} &\cellcolor{gray!25} {\textbf{1.27}} &\cellcolor{gray!25}{\textbf{0.6105}}\\
        \bottomrule
    \end{tabular}}\label{tab:multimodal}
\end{center}
\end{table*}
\begin{figure*}[!t]
 	  \centering
 			\includegraphics[width=1.0\textwidth]{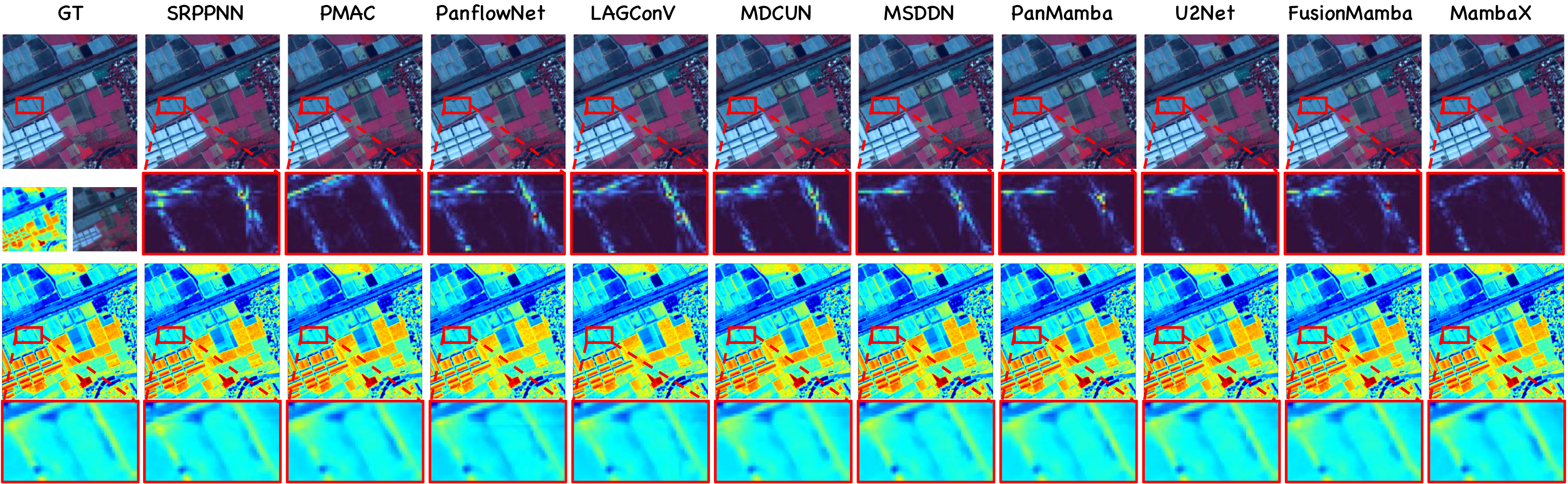}
         \caption{{Visual quality comparison on GF2 datasets. The first and second rows indicate the pseudo-color images of the SR results and the local display of mean squared error (MSE). The third and fourth rows indicate the fourth band of the SR results and their local enlargements.}}\label{gf2}
 \end{figure*}

\noindent\textbf{Quantitative Comparison.} Tab. \ref{Chikusei_Pavia} presents the experimental results on Chikusei and Pavia datasets, demonstrating that MambaX consistently achieves the best results in six experimental settings. MambaX demonstrates the potential advantages of leveraging global semantics through the Mamba framework while effectively integrating spatial and spectral information in state control variable generation. These characteristics may partially explain its consistently strong performance across the three degradation scenarios on both datasets.

\noindent\textbf{Visual Comparison.} Fig. \ref{chikusei} presents pseudo-color images of the reconstructed results for 4× and 8× degraded images from the Chikusei dataset, along with residual maps for the 8× case. For 4× degradation, all methods—except Bicubic—show minimal visual differences from the ground truth. Fig. \ref{pavia} displays the 8× reconstruction results for the Pavia dataset, while Fig. \ref{figall} plots RMSE values across spectral bands. Notably, MambaX consistently achieves the lowest errors, demonstrating its superior spectral curve preservation and effectiveness in restoring long-range spectral information.

\subsection{Multimodal Fusion Super-Resolution}
\subsubsection{Experimental Datasets} 
\textbf{WorldView-III (WV3)}\footnote{https://github.com/liangjiandeng/PanCollection} primarily focuses on capturing urban environments, including roads and cityscapes. The training samples are organized as image triplets in the PAN/LRMS/GT format: the panchromatic (PAN) image has dimensions of 64 × 64 × 1, the low-resolution multispectral (LRMS) image is 16 × 16 × 8, and the ground truth (GT) image is 64 × 64 × 8. In contrast, the reduced-resolution test samples follow a similar triplet format but exhibit larger dimensions of 256 × 256 × 1 for the PAN image, 64 × 64 × 8 for the LRMS image, and 256 × 256 × 8 for the GT image. 

\noindent\textbf{Gaofen-2 (GF2)}\footnote{https://github.com/liangjiandeng/PanCollection} focuses on natural landscapes, including mountainous and riverine regions. Like WV3, it adheres to the Wald protocol for dataset generation and contains 20 reduced-resolution testing samples.

 \begin{figure*}[!t]
 	  \centering
 			\includegraphics[width=1.0\textwidth]{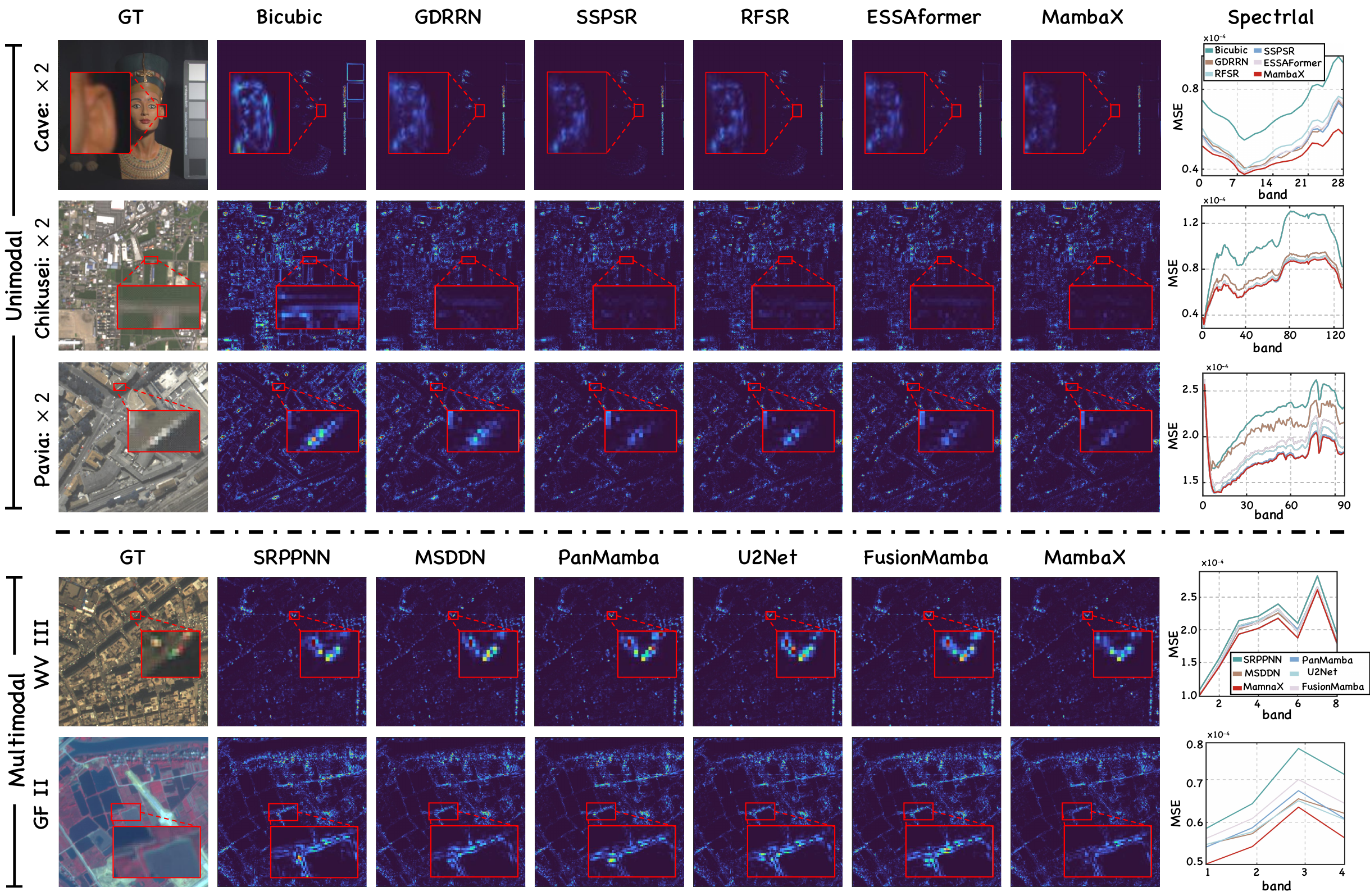}
         \caption{{Visual quality comparison on CAVE, Chikusei, Pavia, WV3 and GF2 datasets. The right curves are the root mean squared error (RMSE) curves for each band between different algorithms.}}
 \label{figall}
 \end{figure*}
 
\subsubsection{Comparison with Professional Methods} 
For the fusion-based SR task, we select a range of representative mainstream algorithms for comparison, including SR-guided progressive pansharpening neural
network (SRPPNN) \cite{srppn}, which uses PAN-guided images for reconstruction at different scales, parallel multiscale attention constraint network (PMAC) \cite{pmac}, which combines multi-scale CNN and transformer, local-context adaptive Conolution kernels-based LAGNet \cite{lag}, multiscale dual-domain guidance network (MSDDN) \cite{msddn}, which uses both spatial and frequency domain guidance, U2Net \cite{u2net} with a dual-stream CNN encoder-decoder structure, flow-Based Deep Network (PanflowNet) \cite{PanFlowNetAF} and memory-augmented deep conditional unfolding network (MDCUN) \cite{MDCUN}, which has strong network interpretability, and the Mamba-based methods PanMamba \cite{HE2025102779} and FusionMamba \cite{FusionMamba}.

\noindent\textbf{Quantitative Comparison.} Tab. \ref{tab:multimodal} presents the quantitative experimental results of various algorithms on two datasets related to fusion-aware SR, i.e., WV3 and GF2. In the WV3 dataset, MambaX outperforms other algorithms, achieving the highest PSNR and SSIM values, with accuracy improvements ranging from 1.53 dB to 0.40 dB and 0.0082 to 0.0024, respectively. Additionally, MambaX achieves the lowest SAM and ERGAS values, indicating superior performance in spectral fidelity and global error control. On the GF2 dataset, MambaX consistently achieves outstanding results in both the WV3 and GF2 scenarios, which further supports the advantage of the dynamically learnable structure in MambaX compared to the original Mamba methods.

\noindent\textbf{Visual Comparison.} From Fig. \ref{gf2}, it is evident that SRPPNN and PanflowNet exhibit noticeable blurring and color distortion in the band images, making it difficult to accurately reproduce spectral details. Additionally, the residual maps reveal that Mamba outperforms other algorithms, especially in the reconstruction of narrow roads and road junctions. Fig. \ref{wv} showcases the results of various algorithms in the building and coastal scenes of the WV dataset. We can observe that Mamba effectively suppresses outliers in the reconstruction of building details, leading to higher-quality images. In the coastal scene, the waves exhibit irregular textures and spatial structures. In Fig. \ref{figall}, we present the performance of MambaX in urban dense areas and agricultural fields. It is evident that, both in terms of overall reconstruction quality and land-cover detail, MambaX delivers superior results compared to mainstream algorithms.

\begin{table*}[!t]
\caption{Ablation Analysis of the proposed MambaX in terms of domain conversion, control matrix learning, and state-level fusion strategy, respectively.}
\centering 
\subtable[Domain Conversion]{        
  {\begin{tabular}{cccc}
	\toprule
	\multirow{3}{*}{Variant} &\multicolumn{3}{c}{Pavia}\\
        \cmidrule(lr){2-4}
        & PSNR & SSIM & SAM \\
        \midrule\midrule
        Bicubic & {35.18} & {94.92} & {3.72} \\
        PixelShuffle & {35.24} & {94.91}  & 3.76 \\
        w/o Weights & {35.37} & 95.03 & 3.70 \\
        \textbf{MambaX} &\textcolor{cyan}{\textbf{35.43}} & \textcolor{cyan}{\textbf{95.16}} &\textcolor{cyan}{\textbf{3.64}} \\
	\bottomrule
	\end{tabular}}
\label{domain}
}
\hfill
\subtable[Contral Matrix Learning]{        
{\begin{tabular}{cccc}
	\toprule
	\multirow{3}{*}{Matrix} &\multicolumn{3}{c}{Gaofen II}\\
        \cmidrule(lr){2-4}
        & PSNR & SSIM & SAM \\
        \midrule\midrule
        w/o $S_{l}(\cdot )$ & {43.52} & {96.37} & {1.47} \\
        w/o $C_{l}(\cdot )$ & {43.50} & {97.39} & 1.32 \\
        linear & 43.12 & 98.29 & 1.53 \\
        \textbf{nSPC} & \textcolor{cyan}{\textbf{43.65}} & \textcolor{cyan}{\textbf{98.38}} &\textcolor{cyan}{\textbf{1.27}} \\
	\bottomrule
	\end{tabular}}
\label{matrix}
}
\hfill
\subtable[State-level Fusion Strategy]{
\begin{tabular}{ccccc}
        \toprule
\multirow{3}{*}{Variant} &\multicolumn{3}{c}{Worldview III}\\
        \cmidrule(lr){2-4}
         & PSNR & SSIM & SAM  \\
        \midrule\midrule
        $\{B, C, \bigtriangleup \} \Rightarrow X $ & 37.34 & 97.31 & 3.93  \\ 
        $\{B, \bigtriangleup\} \Rightarrow X$, \textcolor{magenta}{$\{C\} \Rightarrow S$} & {37.75} & {97.49} &{3.73}  \\
        $\{\bigtriangleup\} \Rightarrow X$, \textcolor{magenta}{$\{B, C\} \Rightarrow S$} & 37.34 & 97.39 & 3.91 \\
        $\{C, \bigtriangleup\} \Rightarrow X$, \textcolor{magenta}{$\{B\} \Rightarrow S$} & \textcolor{cyan}{\textbf{37.87}} & \textcolor{cyan}{\textbf{97.55}} & \textcolor{cyan}{\textbf{3.66}} \\  
	\bottomrule
\end{tabular}
\label{fusion}
}
\end{table*}
\begin{figure*}[!t]
 	  \centering
			\includegraphics[width=1.0\textwidth]{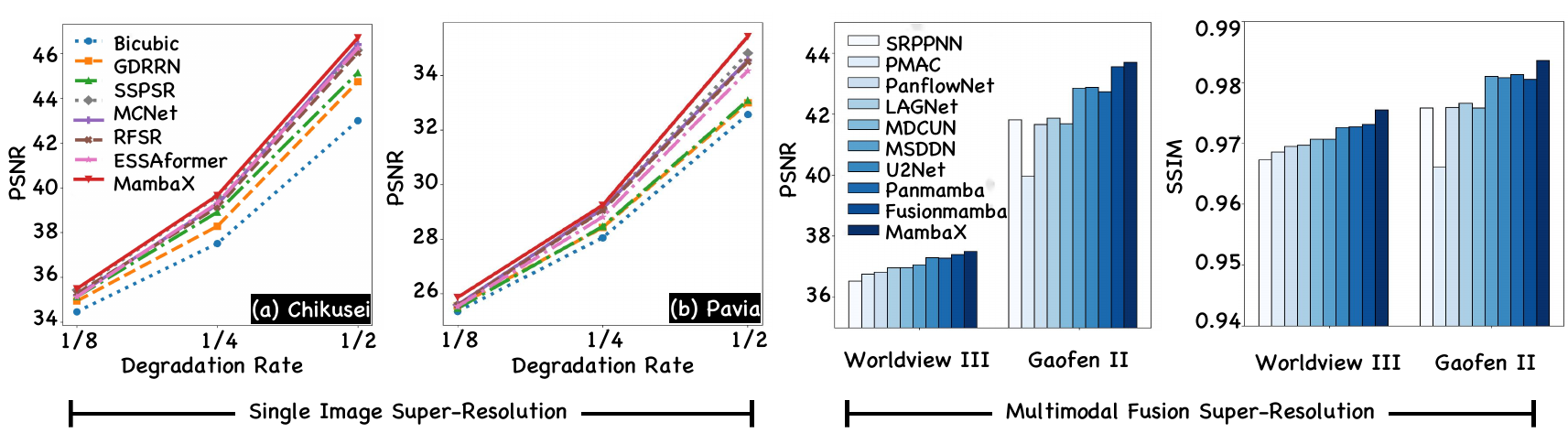}
         \caption{{Performance comparison of MambaX with other SOTA methods on Chikusei, Pavia, WV III, and GF II datasets.}}\label{abl}
\end{figure*}

\subsection{Ablation Study}
\subsubsection{Efficiency of Nonlinear Control Matrix}
To analyze the performance of this module, we examined different state control models and evaluated the impact of spectral and spatial information extraction on the control variable generation process. From Tab. \ref{matrix}, we observe that when the generation of state control variables is entirely dynamic, the performance is inferior to when all variables are generated linearly. This suggests that relying solely on dynamic forms for control variable generation may not necessarily enhance model performance. In the ablation study, we found that in the WV3 dataset, spectral operators play a more significant role, while in the GF2 dataset, spatial operators are more important. This discrepancy can be attributed to the fact that WV3 contains more bands than GF2, which strongly emphasizes the importance of spectral operators in multi-band image super-resolution.

Additionally, we observed that in the WV3 dataset, when all state control variables are generated linearly, the performance is similar to that of MambaX when spatial operators are not considered. In the GF2 dataset, the performance is comparable to that of MambaX when spectral operators are disregarded. However, the complete MambaX structure consistently yields the best performance in both datasets. This indicates that relying solely on linear forms prevents the state-space model from fully considering both spatial and spectral information of the image, further validating the rationale behind our research approach.

\subsubsection{Efficiency of Progressive Domain Mapping}
In this subsection, we compare the impact of different upsampling methods on SR results. The experimental results are presented in Tab. \ref{domain}. We observe that, for 2x and 4x degradation scales, the performance of SR structures based on quadratic spline interpolation and those combining convolution with PixelShuffle is quite similar. However, at 8x degradation scale, the quadratic spline-based method yields better accuracy. This may be because single-scale upsampling methods based on DL struggle to recover high-degradation images due to the limitations of their receptive fields, leading to instability in their performance. On the other hand, bicubic interpolation offers a consistent upsampling operation, which enables it to maintain relatively good performance even on heavily degraded images.

\subsubsection{Efficiency of Cross-control Mechanism State Fusion}
To evaluate different state-level fusion strategies, we assessed the model's performance when generating control variables using either $X$ or auxiliary images $S$ separately. As shown in Tab. \ref{fusion}, generating control variables solely from $X$ results in inferior performance compared to using $S$, highlighting the advantage of auxiliary variables in state-level image fusion. Further analysis reveals that when $S$ generates all control variables, the model outperforms configurations where $S$ generates only $B$ and $C$ in the WV3 dataset. This suggests that the discrete variable $\bigtriangleup$ provides performance gains to some extent. Despite performance variations across datasets, even the least effective fusion strategies achieve competitive results, reinforcing the effectiveness of state-level fusion in SR tasks.

Notably, the model achieves optimal performance when $S$ is used to generate $B$. This is likely due to $B$’s role in modulating $X$, introducing less disruption compared to direct control via $C$ or $\bigtriangleup$. Moreover, the interaction between $B$ and $X$ enables an initial fusion stage that, through subsequent state transitions, better integrates global spatial semantics, enhancing the utilization of auxiliary information.
Comparison results of PSNR and the number of parameters. The performance points of the same method across different datasets are represented in the same color and connected by dashed lines.

\begin{figure}[!t]
 	  \centering
			\includegraphics[width=0.35\textwidth]{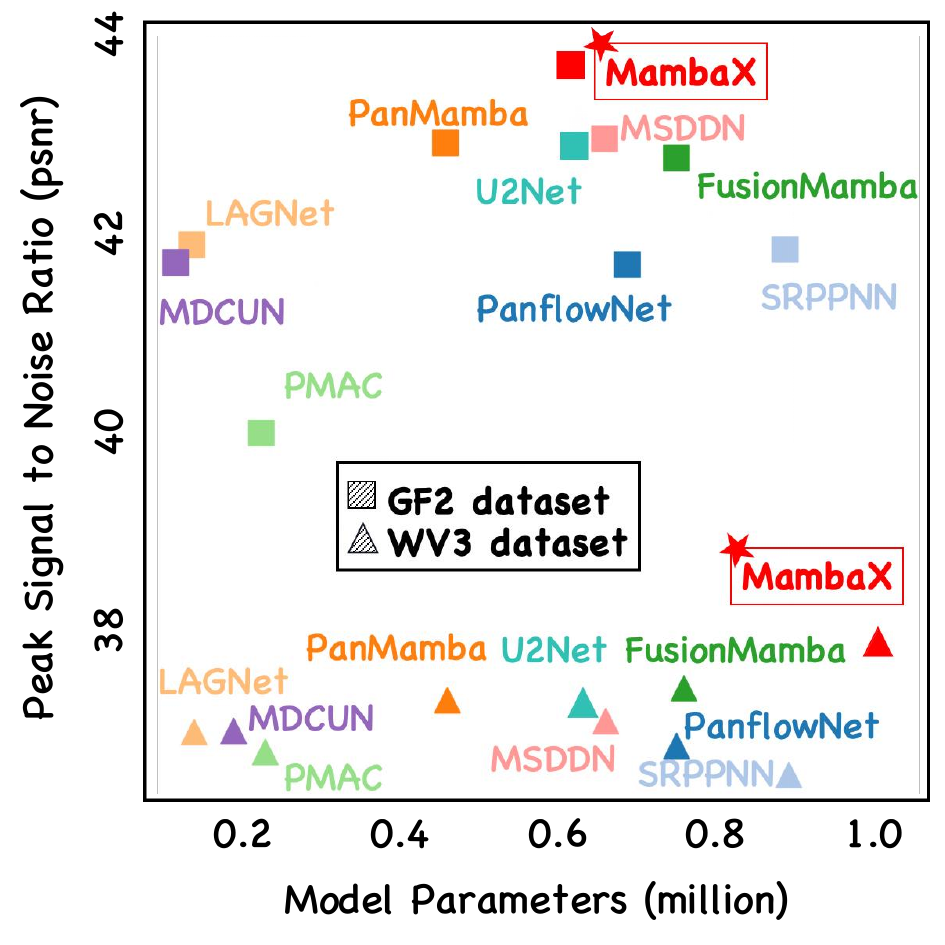}
         \caption{{Comparison of PSNR and the number of parameters. The performance points of the same method across different datasets are represented in the same color and connected by dashed lines.}}\label{para}
\end{figure}
\section{Conclusion}
We propose MambaX, a dynamic learning framework for nonlinear control matrices, which leverages a customized neural network module to adaptively compute control coefficients across multiple stages of the state model. This enables effective regulation and optimization of the blind SR process. Compared to baseline models, MambaX expands the learnable scope of the control matrix within the state model, thereby significantly enhancing the applicability of linearized dynamics in the latent space. Furthermore, to mitigate heterogeneity arising from domain or dimensional discrepancies, we introduce a cross-domain transition operator that progressively reduces error accumulation, ensuring a seamless transformation from the spectral space to the state space.

Furthermore, we are the first to propose a cross-control mechanism for multimodal fusion, wherein the matrix parameters governing the multimodal state properties are leveraged within the latent state space to guide the optimization trajectory of another modality. Empirically, we evaluate our approach across both low- and high-dimensional settings, as well as under unimodal and multimodal conditions for image super-resolution. The results demonstrate that MambaX achieves performance that is on par with state-of-the-art stochastic SSMs. Despite its promising performance, MambaX remains a multi-stage composition of linear systems. To further enhance its adaptability across diverse tasks, we plan to integrate uncertainty analysis by modeling a complex-valued stochastic variable for state representation.

\ifCLASSOPTIONcompsoc
  \section*{Acknowledgments}
\else
  \section*{Acknowledgment}
\fi

This work was supported by the National Natural Science Foundation of China under Grant 42271350 and by the International Partnership Program of the Chinese Academy of Sciences under Grant No.313GJHZ2023066FN.

\begin{figure}[!t]
 	  \centering
			\includegraphics[width=0.5\textwidth]{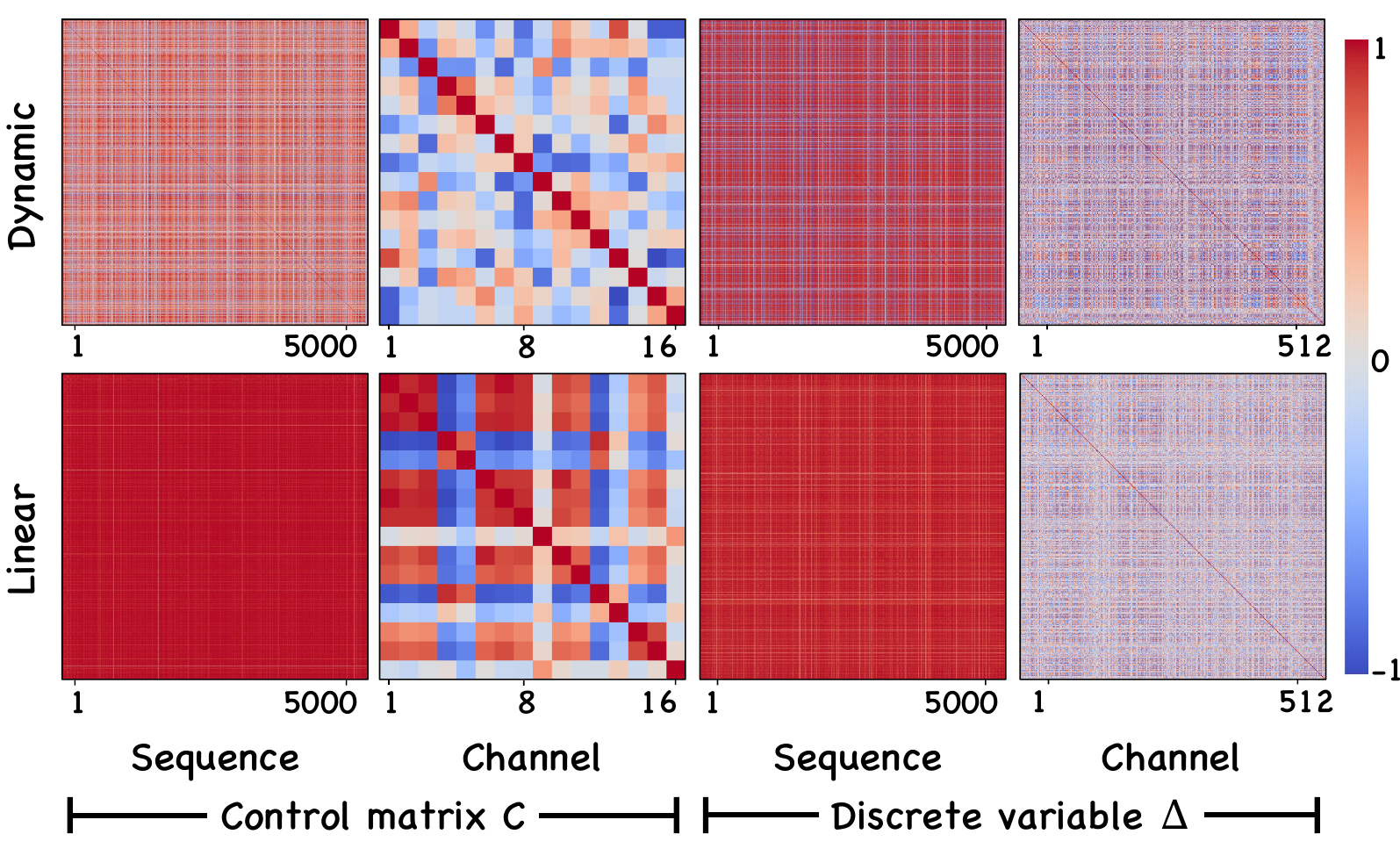}
         \caption{{Comparison results of the Pearson Correlation Coefficient of $\textbf{C}$ and $\bigtriangleup$ generated by linear and dynamic methods in space and channel.}}\label{correlation}
\end{figure}

\bibliographystyle{IEEEtran}
\bibliography{MambaX_ref}

@ARTICLE{blindsuperreview,
  author={Liu, Anran and Liu, Yihao and Gu, Jinjin and Qiao, Yu and Dong, Chao},
  journal={IEEE Transactions on Pattern Analysis and Machine Intelligence}, 
  title={Blind Image Super-Resolution: A Survey and Beyond}, 
  year={2023},
  volume={45},
  number={5},
  pages={5461-5480},
  doi={10.1109/TPAMI.2022.3203009}}

@article{gao2022bayesian,
  title={Bayesian image super-resolution with deep modeling of image statistics},
  author={Gao, Shangqi and Zhuang, Xiahai},
  journal={IEEE Transactions on Pattern Analysis and Machine Intelligence},
  volume={45},
  number={2},
  pages={1405--1423},
  year={2022},
  publisher={IEEE}
}

@ARTICLE{2010Sparse,
  author={Kim, Kwang In and Kwon, Younghee},
  journal={IEEE Transactions on Pattern Analysis and Machine Intelligence}, 
  title={Single-Image Super-Resolution Using Sparse Regression and Natural Image Prior}, 
  year={2010},
  volume={32},
  number={6},
  pages={1127-1133},
  doi={10.1109/TPAMI.2010.25}}

@ARTICLE{2013Dictionaries,
  author={Jia, Kui and Wang, Xiaogang and Tang, Xiaoou},
  journal={IEEE Transactions on Pattern Analysis and Machine Intelligence}, 
  title={Image Transformation Based on Learning Dictionaries across Image Spaces}, 
  year={2013},
  volume={35},
  number={2},
  pages={367-380},
  doi={10.1109/TPAMI.2012.95}}

@ARTICLE{2024Multi-Modal,
  author={Deng, Xin and Xu, Jingyi and Gao, Fangyuan and Sun, Xiancheng and Xu, Mai},
  journal={IEEE Transactions on Pattern Analysis and Machine Intelligence}, 
  title={Deep$\mathrm {M^{2}}$M2CDL: Deep Multi-Scale Multi-Modal Convolutional Dictionary Learning Network}, 
  year={2024},
  volume={46},
  number={5},
  pages={2770-2787},
  doi={10.1109/TPAMI.2023.3334624}}

@ARTICLE{2024Spatial-Frequency,
  author={Zhou, Man and Huang, Jie and Yan, Keyu and Hong, Danfeng and Jia, Xiuping and Chanussot, Jocelyn and Li, Chongyi},
  journal={IEEE Transactions on Pattern Analysis and Machine Intelligence}, 
  title={A General Spatial-Frequency Learning Framework for Multimodal Image Fusion}, 
  year={2024},
  volume={},
  number={},
  pages={1-18},
  doi={10.1109/TPAMI.2024.3368112}}

@ARTICLE{2021Multi-Modal,
  author={Deng, Xin and Dragotti, Pier Luigi},
  journal={IEEE Transactions on Pattern Analysis and Machine Intelligence}, 
  title={Deep Convolutional Neural Network for Multi-Modal Image Restoration and Fusion}, 
  year={2021},
  volume={43},
  number={10},
  pages={3333-3348},
  doi={10.1109/TPAMI.2020.2984244}}

@ARTICLE{MHF-net,
  author={Xie, Qi and Zhou, Minghao and Zhao, Qian and Xu, Zongben and Meng, Deyu},
  journal={IEEE Transactions on Pattern Analysis and Machine Intelligence}, 
  title={MHF-Net: An Interpretable Deep Network for Multispectral and Hyperspectral Image Fusion}, 
  year={2022},
  volume={44},
  number={3},
  pages={1457-1473},
  doi={10.1109/TPAMI.2020.3015691}}

@ARTICLE{2023diffusion,
  author={Saharia, Chitwan and Ho, Jonathan and Chan, William and Salimans, Tim and Fleet, David J. and Norouzi, Mohammad},
  journal={IEEE Transactions on Pattern Analysis and Machine Intelligence}, 
  title={Image Super-Resolution via Iterative Refinement}, 
  year={2023},
  volume={45},
  number={4},
  pages={4713-4726},
  doi={10.1109/TPAMI.2022.3204461}}

@ARTICLE{2015CNN,
  author={Dong, Chao and Loy, Chen Change and He, Kaiming and Tang, Xiaoou},
  journal={IEEE Transactions on Pattern Analysis and Machine Intelligence}, 
  title={Image Super-Resolution Using Deep Convolutional Networks}, 
  year={2016},
  volume={38},
  number={2},
  pages={295-307},
  doi={10.1109/TPAMI.2015.2439281}}

@ARTICLE{2022lap,
  author={Anwar, Saeed and Barnes, Nick},
  journal={IEEE Transactions on Pattern Analysis and Machine Intelligence}, 
  title={Densely Residual Laplacian Super-Resolution}, 
  year={2022},
  volume={44},
  number={3},
  pages={1192-1204},
  doi={10.1109/TPAMI.2020.3021088}}

@Inbook{Grancharova2012,
author="Grancharova, Alexandra
and Johansen, Tor Arne",
title="Nonlinear Model Predictive Control",
bookTitle="Explicit Nonlinear Model Predictive Control: Theory and Applications",
year="2012",
publisher="Springer Berlin Heidelberg",
address="Berlin, Heidelberg",
pages="39--69",
isbn="978-3-642-28780-0",
doi="10.1007/978-3-642-28780-0_2",
url="https://doi.org/10.1007/978-3-642-28780-0_2"
}

@misc{ssmlayerwisenonlinearity,
      title={State-space Models with Layer-wise Nonlinearity are Universal Approximators with Exponential Decaying Memory}, 
      author={Shida Wang and Beichen Xue},
      year={2023},
      eprint={2309.13414},
      archivePrefix={arXiv},
      primaryClass={cs.LG},
      url={https://arxiv.org/abs/2309.13414}, 
}

@misc{mathematicalssm,
      title={Mathematical Formalism for Memory Compression in Selective State Space Models}, 
      author={Siddhanth Bhat},
      year={2024},
      eprint={2410.03158},
      archivePrefix={arXiv},
      primaryClass={cs.LG},
      url={https://arxiv.org/abs/2410.03158}, 
}

@misc{gu2024mambalineartimesequencemodeling,
      title={Mamba: Linear-Time Sequence Modeling with Selective State Spaces}, 
      author={Albert Gu and Tri Dao},
      year={2024},
      eprint={2312.00752},
      archivePrefix={arXiv},
      primaryClass={cs.LG},
      url={https://arxiv.org/abs/2312.00752}, 
}

@ARTICLE{CAVE,
  author={Yasuma, Fumihito and Mitsunaga, Tomoo and Iso, Daisuke and Nayar, Shree K.},
  journal={IEEE Transactions on Image Processing}, 
  title={Generalized Assorted Pixel Camera: Postcapture Control of Resolution, Dynamic Range, and Spectrum}, 
  year={2010},
  volume={19},
  number={9},
  pages={2241-2253}}

@misc{adam,
      title={Adam: A Method for Stochastic Optimization}, 
      author={Diederik P. Kingma and Jimmy Ba},
      year={2017},
      eprint={1412.6980},
      archivePrefix={arXiv},
      primaryClass={cs.LG},
      url={https://arxiv.org/abs/1412.6980}, 
}

@inproceedings{u2net,
author = {Peng, Siran and Guo, Chenhao and Wu, Xiao and Deng, Liang-Jian},
title = {U2Net: A General Framework with Spatial-Spectral-Integrated Double U-Net for Image Fusion},
year = {2023},
isbn = {9798400701085},
publisher = {Association for Computing Machinery},
address = {New York, NY, USA},
doi = {10.1145/3581783.3612084},
booktitle = {Proceedings of the 31st ACM International Conference on Multimedia},
pages = {3219–3227},
numpages = {9},
location = {Ottawa ON, Canada},
series = {MM '23}
}

@ARTICLE{msddn,
  author={He, Xuanhua and Yan, Keyu and Zhang, Jie and Li, Rui and Xie, Chengjun and Zhou, Man and Hong, Danfeng},
  journal={IEEE Transactions on Geoscience and Remote Sensing}, 
  title={Multiscale Dual-Domain Guidance Network for Pan-Sharpening}, 
  year={2023},
  volume={61},
  number={},
  pages={1-13},
  doi={10.1109/TGRS.2023.3273334}}

@article{PanFlowNetAF,
  title={PanFlowNet: A Flow-Based Deep Network for Pan-sharpening},
  author={Gang Yang and Xiangyong Cao and Wen Xiao and Man Zhou and Aiping Liu and Xun Chen and Deyu Meng},
  journal={2023 IEEE/CVF International Conference on Computer Vision (ICCV)},
  year={2023},
  pages={16811-16821}
}

@ARTICLE{srppn,
  author={Cai, Jiajun and Huang, Bo},
  journal={IEEE Transactions on Geoscience and Remote Sensing}, 
  title={Super-Resolution-Guided Progressive Pansharpening Based on a Deep Convolutional Neural Network}, 
  year={2021},
  volume={59},
  number={6},
  pages={5206-5220},
  doi={10.1109/TGRS.2020.3015878}}

@INPROCEEDINGS{MDCUN,
  author={Yang, Gang and Zhou, Man and Yan, Keyu and Liu, Aiping and Fu, Xueyang and Wang, Fan},
  booktitle={2022 IEEE/CVF Conference on Computer Vision and Pattern Recognition (CVPR)}, 
  title={Memory-augmented Deep Conditional Unfolding Network for Pansharpening}, 
  year={2022},
  volume={},
  number={},
  pages={1778-1787},
  doi={10.1109/CVPR52688.2022.00183}}

@ARTICLE{pmac,
  author={Liang, Yixun and Zhang, Ping and Mei, Yang and Wang, Tingqi},
  journal={IEEE Geoscience and Remote Sensing Letters}, 
  title={PMACNet: Parallel Multiscale Attention Constraint Network for Pan-Sharpening}, 
  year={2022},
  volume={19},
  number={},
  pages={1-5},
  doi={10.1109/LGRS.2022.3170904}}

@article{lag, title={LAGConv: Local-Context Adaptive Convolution Kernels with Global Harmonic Bias for Pansharpening}, volume={36}, DOI={10.1609/aaai.v36i1.19996}, number={1}, journal={Proceedings of the AAAI Conference on Artificial Intelligence}, author={Jin, Zi-Rong and Zhang, Tian-Jing and Jiang, Tai-Xiang and Vivone, Gemine and Deng, Liang-Jian}, year={2022}, month={Jun.}, pages={1113-1121} }

@article{essaformer,
  title={ESSAformer: Efficient Transformer for Hyperspectral Image Super-resolution},
  author={Mingjin Zhang and Chi Zhang and Qiming Zhang and Jie-Ru Guo and Xinbo Gao and Jing Zhang},
  journal={2023 IEEE/CVF International Conference on Computer Vision (ICCV)},
  year={2023},
  pages={23016-23027}
}

@ARTICLE{rfsr,
  author={Wang, Xinya and Ma, Jiayi and Jiang, Junjun},
  journal={IEEE Transactions on Geoscience and Remote Sensing}, 
  title={Hyperspectral Image Super-Resolution via Recurrent Feedback Embedding and Spatial–Spectral Consistency Regularization}, 
  year={2022},
  volume={60},
  number={},
  pages={1-13},
  doi={10.1109/TGRS.2021.3064450}}

@ARTICLE{sspsr,
  author={Jiang, Junjun and Sun, He and Liu, Xianming and Ma, Jiayi},
  journal={IEEE Transactions on Computational Imaging}, 
  title={Learning Spatial-Spectral Prior for Super-Resolution of Hyperspectral Imagery}, 
  year={2020},
  volume={6},
  number={},
  pages={1082-1096},
  doi={10.1109/TCI.2020.2996075}}

@INPROCEEDINGS{GDRRN,
  author={Li, Yong and Zhang, Lei and Dingl, Chen and Wei, Wei and Zhang, Yanning},
  booktitle={2018 IEEE Fourth International Conference on Multimedia Big Data (BigMM)}, 
  title={Single Hyperspectral Image Super-Resolution with Grouped Deep Recursive Residual Network}, 
  year={2018},
  volume={},
  number={},
  pages={1-4},
  doi={10.1109/BigMM.2018.8499097}}

@Article{MCNET,
AUTHOR = {Li, Qiang and Wang, Qi and Li, Xuelong},
TITLE = {Mixed 2D/3D Convolutional Network for Hyperspectral Image Super-Resolution},
JOURNAL = {Remote Sensing},
VOLUME = {12},
YEAR = {2020},
NUMBER = {10},
ARTICLE-NUMBER = {1660},
ISSN = {2072-4292}}

@ARTICLE{3dqrnn,
  author={Fu, Ying and Liang, Zhiyuan and You, Shaodi},
  journal={IEEE Journal of Selected Topics in Applied Earth Observations and Remote Sensing}, 
  title={Bidirectional 3D Quasi-Recurrent Neural Network for Hyperspectral Image Super-Resolution}, 
  year={2021},
  volume={14},
  number={},
  pages={2674-2688},
  doi={10.1109/JSTARS.2021.3057936}}

@misc{gu2020hippo,
      title={HiPPO: Recurrent Memory with Optimal Polynomial Projections}, 
      author={Albert Gu and Tri Dao and Stefano Ermon and Atri Rudra and Christopher Re},
      year={2020},
      eprint={2008.07669},
      archivePrefix={arXiv},
      primaryClass={cs.LG},
      url={https://arxiv.org/abs/2008.07669}, 
}

@article{FusionMamba,
   title={FusionMamba: Efficient Remote Sensing Image Fusion With State Space Model},
   volume={62},
   ISSN={1558-0644},
   DOI={10.1109/tgrs.2024.3496073},
   journal={IEEE Transactions on Geoscience and Remote Sensing},
   publisher={Institute of Electrical and Electronics Engineers (IEEE)},
   author={Peng, Siran and Zhu, Xiangyu and Deng, Haoyu and Deng, Liang-Jian and Lei, Zhen},
   year={2024},
   pages={1–16} }

@article{HE2025102779,
title = {Pan-Mamba: Effective pan-sharpening with state space model},
journal = {Information Fusion},
volume = {115},
pages = {102779},
year = {2025},
issn = {1566-2535},
doi = {https://doi.org/10.1016/j.inffus.2024.102779},
author = {Xuanhua He and Ke Cao and Jie Zhang and Keyu Yan and Yingying Wang and Rui Li and Chengjun Xie and Danfeng Hong and Man Zhou}
}

@ARTICLE{transformer_sup_token,
  author={Xiao, Yi and Yuan, Qiangqiang and Jiang, Kui and He, Jiang and Lin, Chia-Wen and Zhang, Liangpei},
  journal={IEEE Transactions on Image Processing}, 
  title={TTST: A Top-k Token Selective Transformer for Remote Sensing Image Super-Resolution}, 
  year={2024},
  volume={33},
  number={},
  pages={738-752},
  doi={10.1109/TIP.2023.3349004}}

@ARTICLE{CNN_sup_comple,
  author={Zamir, Syed Waqas and Arora, Aditya and Khan, Salman and Hayat, Munawar and Khan, Fahad Shahbaz and Yang, Ming-Hsuan and Shao, Ling},
  journal={IEEE Transactions on Pattern Analysis and Machine Intelligence}, 
  title={Learning Enriched Features for Fast Image Restoration and Enhancement}, 
  year={2023},
  volume={45},
  number={2},
  pages={1934-1948},
  doi={10.1109/TPAMI.2022.3167175}}

@ARTICLE{ssfusion_tra_tensor,
  author={Ye, Fei and Wu, Zebin and Jia, Xiuping and Chanussot, Jocelyn and Xu, Yang and Wei, Zhihui},
  journal={IEEE Transactions on Image Processing}, 
  title={Bayesian Nonlocal Patch Tensor Factorization for Hyperspectral Image Super-Resolution}, 
  year={2023},
  volume={32},
  number={},
  pages={5877-5892},
  doi={10.1109/TIP.2023.3326687}}

@ARTICLE{unrolling_sup_tensor,
  author={Wang, Kaidong and Liao, Xiuwu and Li, Jun and Meng, Deyu and Wang, Yao},
  journal={IEEE Transactions on Image Processing}, 
  title={Hyperspectral Image Super-Resolution via Knowledge-Driven Deep Unrolling and Transformer Embedded Convolutional Recurrent Neural Network}, 
  year={2023},
  volume={32},
  number={},
  pages={4581-4594},
  doi={10.1109/TIP.2023.3293768}}

@article{dosovitskiy2020vit,
  title={An Image is Worth 16x16 Words: Transformers for Image Recognition at Scale},
  author={Dosovitskiy, Alexey and Beyer, Lucas and Kolesnikov, Alexander and Weissenborn, Dirk and Zhai, Xiaohua and Unterthiner, Thomas and  Dehghani, Mostafa and Minderer, Matthias and Heigold, Georg and Gelly, Sylvain and Uszkoreit, Jakob and Houlsby, Neil},
  journal={ICLR},
  year={2021}
}

@ARTICLE{lichengyu2023lowrank,
  author={Li, Chenyu and Zhang, Bing and Hong, Danfeng and Yao, Jing and Chanussot, Jocelyn},
  journal={IEEE Transactions on Geoscience and Remote Sensing}, 
  title={LRR-Net: An Interpretable Deep Unfolding Network for Hyperspectral Anomaly Detection}, 
  year={2023},
  volume={61},
  number={},
  pages={1-12},
  doi={10.1109/TGRS.2023.3279834}}

@ARTICLE{hongdanfeng2023srmapping,
  author={Hong, Danfeng and Yao, Jing and Li, Chenyu and Meng, Deyu and Yokoya, Naoto and Chanussot, Jocelyn},
  journal={IEEE Transactions on Geoscience and Remote Sensing}, 
  title={Decoupled-and-Coupled Networks: Self-Supervised Hyperspectral Image Super-Resolution With Subpixel Fusion}, 
  year={2023},
  volume={61},
  number={},
  pages={1-12},
  doi={10.1109/TGRS.2023.3324497}}

@ARTICLE{lichengyu2024learn-prior,
  author={Li, Chenyu and Zhang, Bing and Hong, Danfeng and Jia, Xiuping and Plaza, Antonio and Chanussot, Jocelyn},
  journal={IEEE Transactions on Neural Networks and Learning Systems}, 
  title={Learning Disentangled Priors for Hyperspectral Anomaly Detection: A Coupling Model-Driven and Data-Driven Paradigm}, 
  year={2024},
  volume={},
  number={},
  pages={1-14},
  doi={10.1109/TNNLS.2024.3401589}}

@article{LI2024CasFormer,
title = {CasFormer: Cascaded transformers for fusion-aware computational hyperspectral imaging},
journal = {Information Fusion},
volume = {108},
pages = {102408},
year = {2024},
issn = {1566-2535},
author = {Chenyu Li and Bing Zhang and Danfeng Hong and Jun Zhou and Gemine Vivone and Shutao Li and Jocelyn Chanussot},
}

@ARTICLE{10812905,
  author={Pan, Zhaojie and Li, Chenyu and Plaza, Antonio and Chanussot, Jocelyn and Hong, Danfeng},
  journal={IEEE Transactions on Geoscience and Remote Sensing}, 
  title={Hyperspectral Image Classification With Mamba}, 
  year={2025},
  volume={63},
  number={},
  pages={1-14},
  doi={10.1109/TGRS.2024.3521411}}

@InProceedings{KXNet,
author="Fu, Jiahong
and Wang, Hong
and Xie, Qi
and Zhao, Qian
and Meng, Deyu
and Xu, Zongben",
editor="Avidan, Shai
and Brostow, Gabriel
and Ciss{\'e}, Moustapha
and Farinella, Giovanni Maria
and Hassner, Tal",
title="KXNet: A Model-Driven Deep Neural Network for Blind Super-Resolution",
booktitle="Computer Vision -- ECCV 2022",
year="2022",
publisher="Springer Nature Switzerland",
address="Cham",
pages="235--253",
isbn="978-3-031-19800-7"
}

\begin{IEEEbiography}[{\includegraphics[width=1in,height=1.25in,clip,keepaspectratio]
{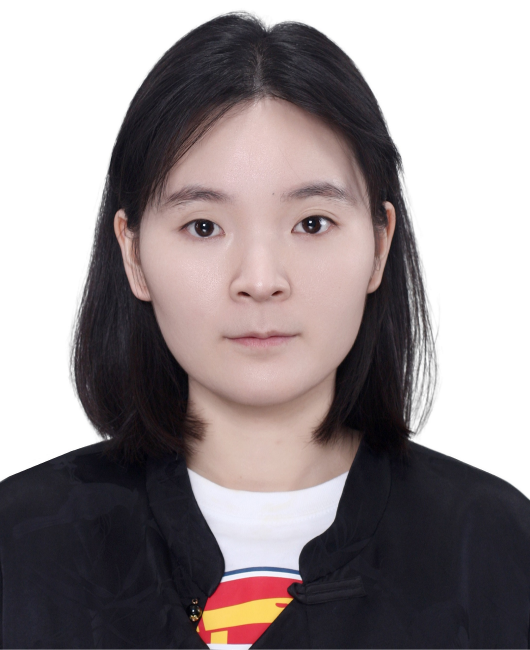}}]{Chenyu Li} received the B.S. and M.S. degrees from the School of Transportation, Southeast University, Nanjing, China, in 2018 and 2021, respectively. She is currently pursuing her Ph.D. degree in mathematics at Southeast University, Nanjing, China. She is also a joint Ph.D. student at the Aerospace Information Research Institute, Chinese Academy of Sciences, Beijing, China. Her research interests include interpretable artificial intelligence, big Earth data forecasting, foundation models, and hyperspectral imaging.
\end{IEEEbiography}

\vskip -2\baselineskip plus -1fil

\begin{IEEEbiography}[{\includegraphics[width=1in,height=1.25in,clip,keepaspectratio]{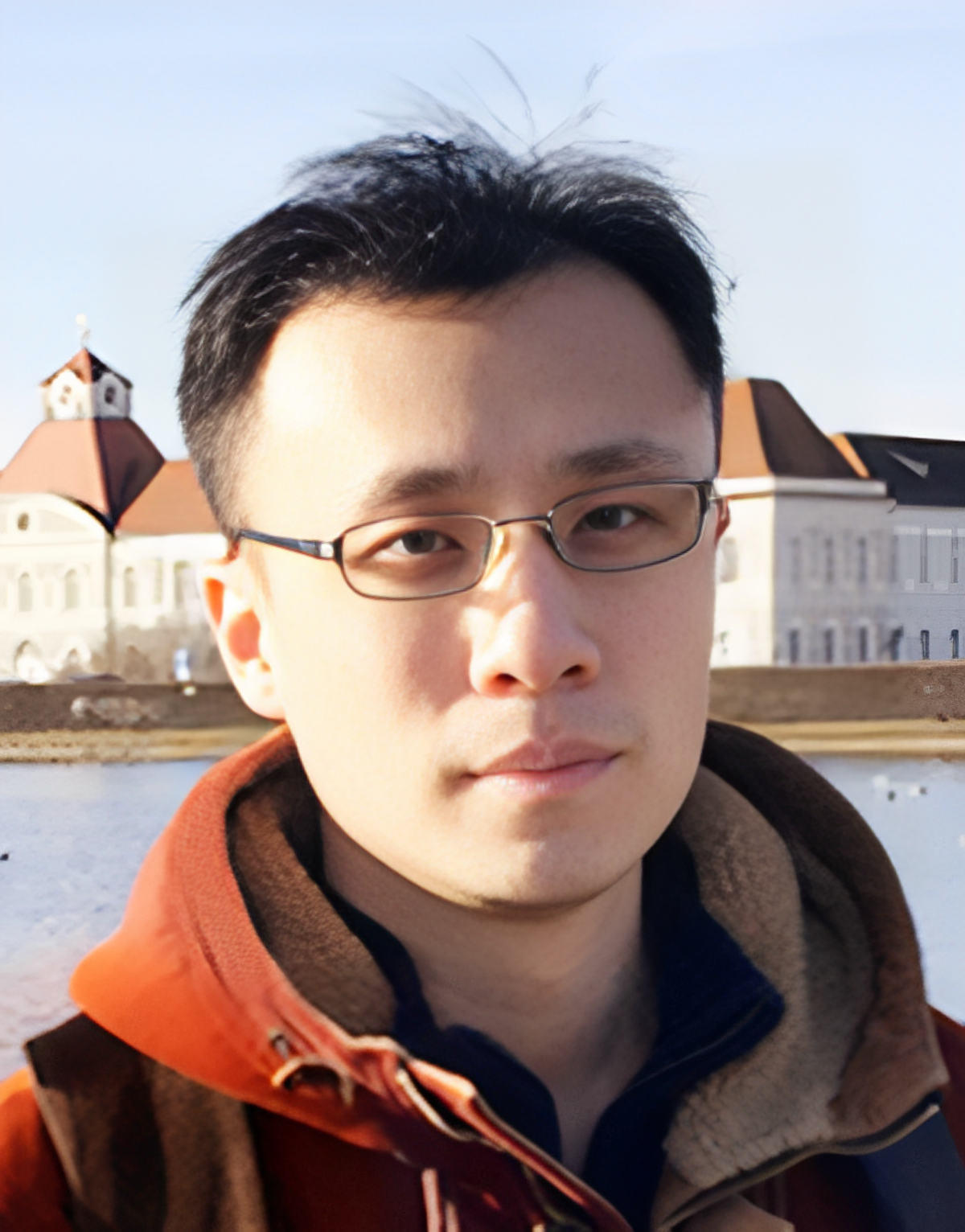}}]{Danfeng Hong}(IEEE Senior Member) received the Dr. -Ing degree (summa cum laude) from the Signal Processing in Earth Observation (SiPEO), Technical University of Munich (TUM), Munich, Germany, in 2019. He is currently a Full Professor at Southeast University, Nanjing, China, and previously served as a Full Professor at the Aerospace Information Research Institute, Chinese Academy of Sciences. His research interests include artificial intelligence, multimodal perception, foundation models, hyperspectral remote sensing, and Earth observation.

Dr. Hong serves as an Associate Editor for the IEEE Transactions on Pattern Analysis and Machine Intelligence (TPAMI), IEEE Transactions on Image Processing (TIP), and IEEE Transactions on Geoscience and Remote Sensing (TGRS). He has been recognized as a Highly Cited Researcher by Clarivate Analytics since 2022.
\end{IEEEbiography}

\vskip -2\baselineskip plus -1fil

\begin{IEEEbiography}[{\includegraphics[width=1in,height=1.25in,clip,keepaspectratio]{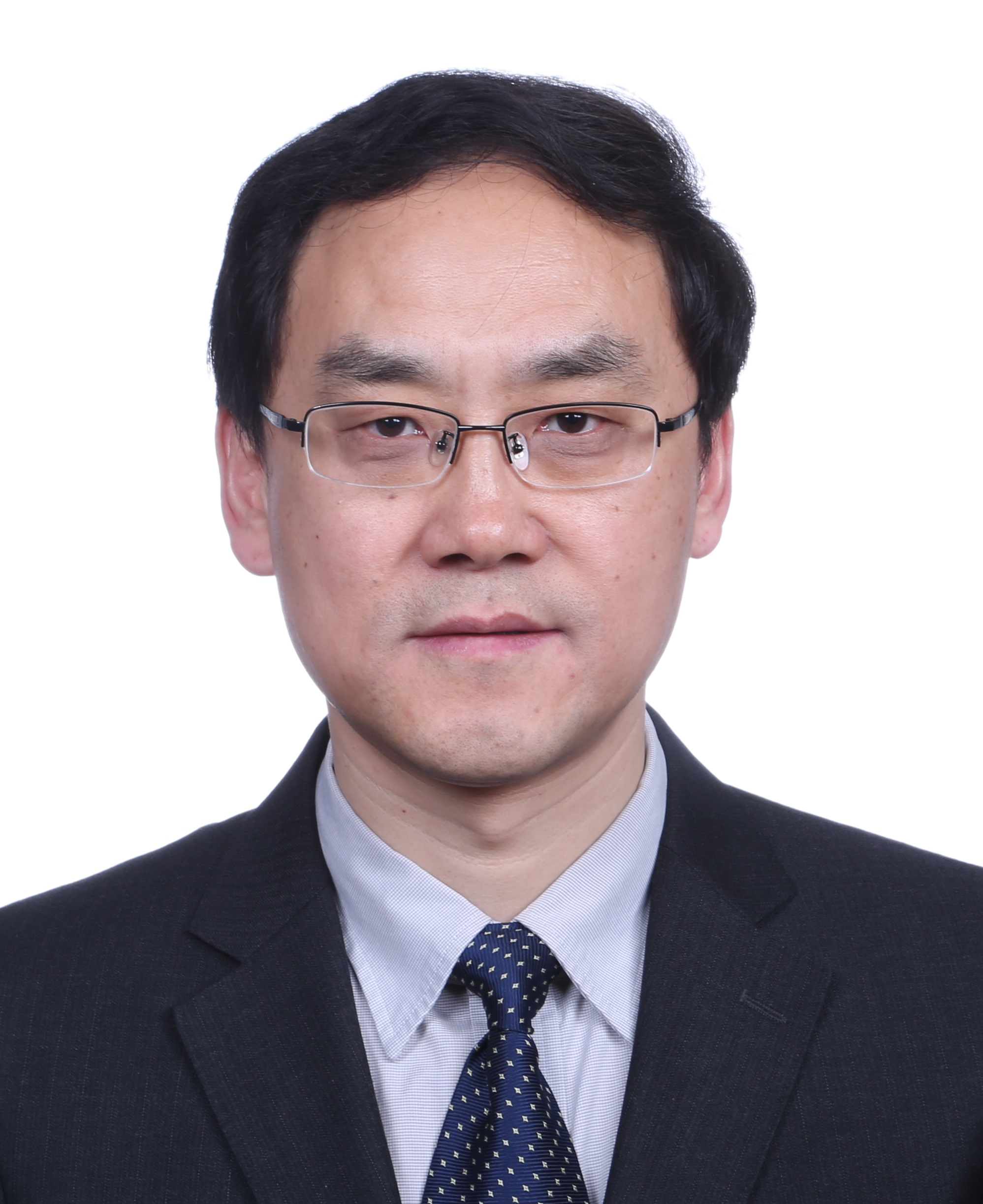}}]{Bing Zhang}(IEEE Fellow) received the B.S. degree in geography from Peking University, Beijing, China, in 1991, and the M.S. and Ph.D. degrees in remote sensing from the Institute of Remote Sensing Applications, Chinese Academy of Sciences, Beijing, in 1994 and 2003, respectively. He is currently a Full Professor and the Deputy Director of the Aerospace Information Research Institute, China Academy of Sciences. He has long been engaged in research on Hyperspectral remote sensing technology and applications. His creative achievements were rewarded with more than 10 important prizes, including the IEEE Geoscience and Remote Sensing Society (GRSS) Regional Leader Award, the National Science and Technology Advance Award of China, the Outstanding Scientific Achievement Award of the Chinese Academy of Sciences, etc.
\end{IEEEbiography}

\vskip -2\baselineskip plus -1fil

\begin{IEEEbiography}[{\includegraphics[width=1in,height=1.25in,clip,keepaspectratio]
{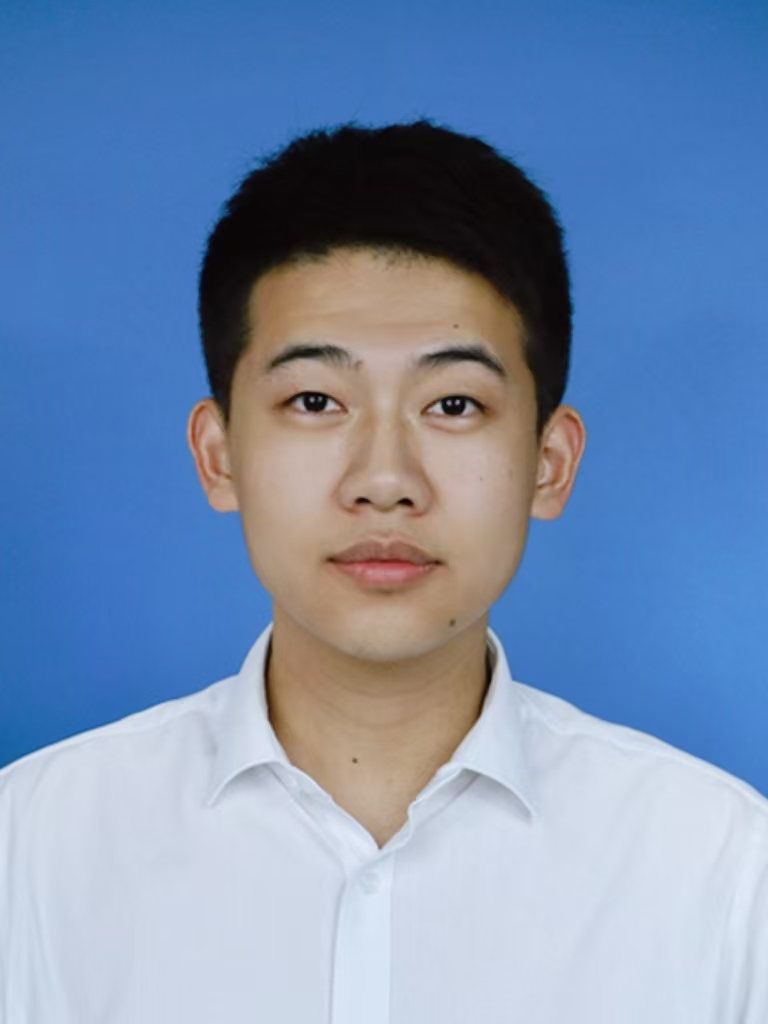}}]{Zhaojie Pan} received the B.S. and M.c. degrees from China University of Petroleum (East China), Qingdao, China, in 2021 and 2024, respectively. He is currently pursuing a Ph.D. degree with the School of Mathematics, Southeast University, Nanjing, China, and is also a joint Ph.D. student at the Aerospace Information Research Institute, Chinese Academy of Sciences, Beijing, China. His research interests include deep learning and remote sensing image processing. 
\end{IEEEbiography}

\vskip -2\baselineskip plus -1fil

\begin{IEEEbiography}[{\includegraphics[width=1in,height=1.25in,clip,keepaspectratio]
{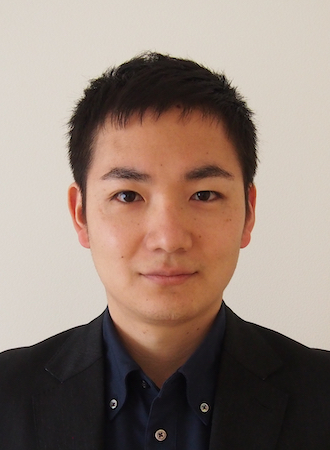}}]{Naoto Yokoya} received the M.Eng. and Ph.D. degrees from the Department of Aeronautics and Astronautics, The University of Tokyo, Tokyo, Japan, in 2010 and 2013, respectively. He is currently an Associate Professor at the University of Tokyo and a Team Leader at the RIKEN Center for Advanced Intelligence Project, RIKEN, Tokyo, where he leads the Geoinformatics Team. His research focuses on image processing, data fusion, and machine learning for understanding remote sensing images, with applications to disaster management and environmental monitoring.

Dr. Yokoya is an Associate Editor of the IEEE Transactions on Pattern Analysis and Machine Intelligence (TPAMI), the IEEE Transactions on Geoscience and Remote Sensing (TGRS), and the ISPRS Journal of Photogrammetry and Remote Sensing. He has been designated a Clarivate Highly Cited Researcher since 2022.
\end{IEEEbiography}

\vskip -2\baselineskip plus -1fil

\begin{IEEEbiography}[{\includegraphics[width=1in,height=1.25in,clip,keepaspectratio]{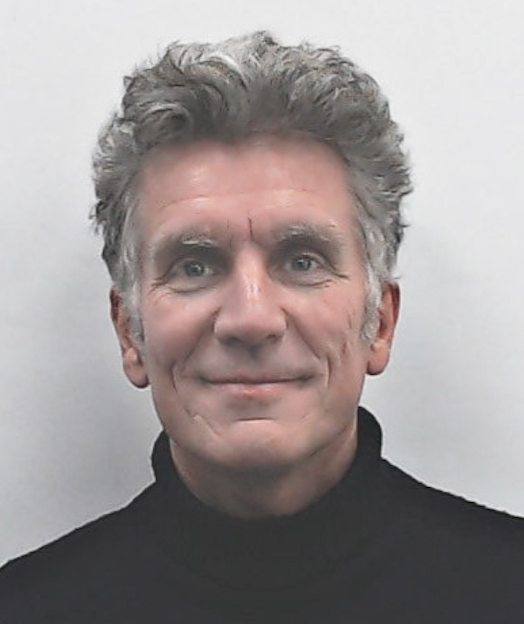}}]{Jocelyn Chanussot} (IEEE Fellow) received the M.Sc. degree in electrical engineering from the Grenoble Institute of Technology (Grenoble INP), Grenoble, France, in 1995, and the Ph.D. degree from the Université de Savoie, Annecy, France, in 1998. From 1999 to 2023, he was with Grenoble INP, where he was a Professor of signal and image processing. He is currently a Research Director with INRIA, Grenoble. His research interests include image analysis, hyperspectral remote sensing, data fusion, machine learning, and artificial intelligence.

Dr. Chanussot is the founding President of the IEEE Geoscience and Remote Sensing French chapter. He was the vice president of the IEEE Geoscience and Remote Sensing Society, in charge of meetings and symposia. He is an Associate Editor for the IEEE Transactions on Geoscience and Remote Sensing, the IEEE Transactions on Image Processing, and the Proceedings of the IEEE. He was the Editor-in-Chief of the IEEE Journal of Selected Topics in Applied Earth Observations and Remote Sensing (2011-2015). He is a Fellow of the IEEE, an ELLIS Fellow, a Fellow of AAIA, a member of the Institut Universitaire de France (2012-2017), and a Highly Cited Researcher (Clarivate Analytics/Thomson Reuters, since 2018).
\end{IEEEbiography}

\end{document}